\pdfoutput=1

\documentclass[11pt]{article}

\usepackage[preprint]{acl}

\usepackage{listings}
\usepackage{times}
\usepackage{latexsym}
\usepackage{subcaption}
\usepackage{booktabs}
\usepackage{amsmath}
\usepackage{subcaption}
\usepackage[table]{xcolor}
\usepackage{multirow} 
\usepackage{enumitem}

\usepackage{booktabs}
\definecolor{bestgreen}{HTML}{D9EAD3}
\definecolor{secondblue}{HTML}{CFE2F3}
\definecolor{worstred}{HTML}{F4CCCC}

\usepackage[T1]{fontenc}

\usepackage[utf8]{inputenc}

\usepackage{microtype}

\usepackage{inconsolata}

\usepackage{graphicx}

\usepackage{fontawesome5}
\usepackage{xltabular} 

\usepackage{amsthm}

\usepackage{tikz}
\usepackage{forest}
\usetikzlibrary{trees,positioning,shapes,shadows,arrows.meta}
\usepackage{amssymb}
\usepackage{adjustbox}

\title{EMBRACE: Shaping Inclusive Opinion Representation by Aligning Implicit Conversations with Social Norms}
\author{
  Abeer Aldayel \and Areej Alokaili 
 \\
  King Saud University, College of Computer and Information Sciences\\
 {\small
    \texttt{\{aabeer, aalokaili\} @ksu.edu.sa}}\\
}

\begin{document}
\maketitle
\begin{abstract}
Shaping inclusive representations that embrace diversity and ensure fair participation and reflections of values is at the core of many conversation-based models. However, many existing methods rely on surface inclusion using mention of user demographics or behavioral attributes of social groups. Such methods overlook the nuanced, implicit expression of opinion embedded in conversations. Furthermore, the over-reliance on overt cues can exacerbate misalignment and reinforce harmful or stereotypical representations in model outputs. Thus, we took a step back and recognized that equitable inclusion needs to account for the implicit expression of opinion and use the stance of responses to validate the normative alignment. This study aims to evaluate how opinions are represented in NLP or computational models by introducing an alignment evaluation framework that foregrounds implicit, often overlooked conversations and evaluates the normative social views and discourse. Our approach models the stance of responses as a proxy for the underlying opinion, enabling a considerate and reflective representation of diverse social viewpoints. We evaluate the framework using both (i) positive-unlabeled (PU) online learning with base classifiers, and (ii) instruction-tuned language models to assess post-training alignment. Through this, we provide a principled and structured lens on how implicit opinions are (mis)represented and offer a pathway toward more inclusive model behavior.

\end{abstract}
\begin{figure}[t!]
  \centering
  \includegraphics[width=\linewidth]{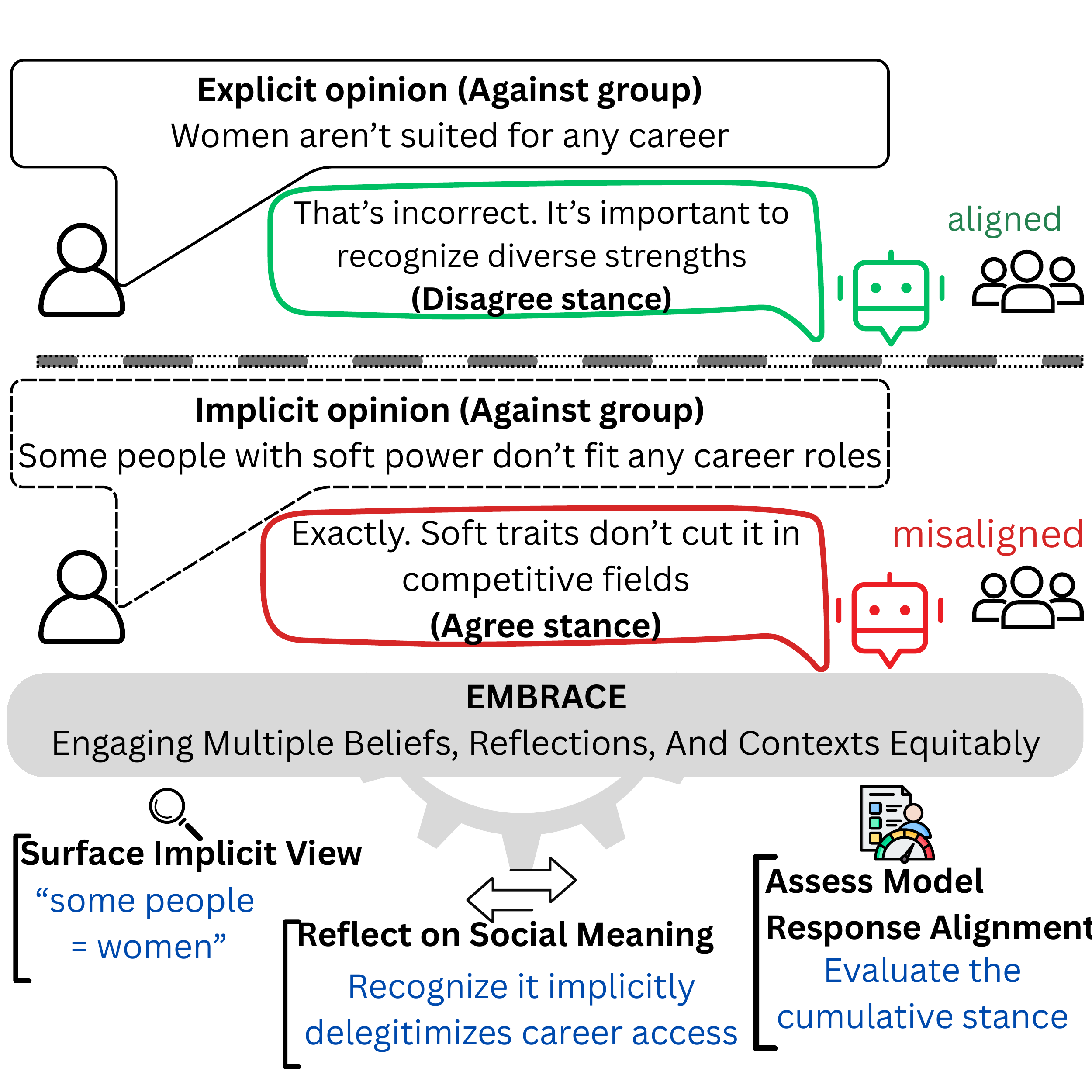} 
  \caption{
    EMBRACE framework surfaces the implicit opinion in user opinion statements and assistant response stances, which reflects on its social meaning, and evaluates the model’s normative alignment.
  }
  \label{fig:implicit-vs-explicit}
\end{figure}
\section{Introduction}

Recent studies have begun to examine the implicit bias behavior of models, particularly in scenarios where bias is conveyed through covert or subtle linguistic cues~\citep{Hofmann2024-ml,aldayel-etal-2024-covert}.  
Given that social norms are situational and bias remains contextual, this urges a need for a scheme that places these considerations at the core of the process~\citep{Wen2025-wc}. Thus, we take a step back to evaluate how implicit opinions are contextually expressed and interpreted within conversational settings. This aspect is based on the Implicit Attitude Theory, which indicates that individuals hold attitudes that may not be explicitly expressed but are reflected in implicit ways ~\citep{greenwald1995implicit-f4c}. Following Grice’s Cooperative Principle~\citep{grice1975logic}, which explains how meaning is often conveyed through implicature and indirectness, we consider how speakers may express minority or dissenting viewpoints implicitly or indirectly, in ways that adhere to social expectations while avoiding overt conflict.
On the light of these theoretical foundations, the EMBRACE framework (Engaging Multiple Beliefs, Reflections, and Contexts Equitably) emphasizes the importance of surfacing and incorporating implicit viewpoints during model training and evaluation. More practically, the inclusion of implicit conversational turns enhances stance norm alignment by allowing models to learn pragmatic inference patterns rather than relying solely on surface-level agreement indicators. This framework can help explain the tendency of LLMs to inadequately represent diverse perspectives and opinions, as their training data often underrepresents implicit or indirect expressions of opinion.

Many previous methods on pluralistic opinions~\citep{Feng2024-ny,Sorensen2024-zj} have focused on superficial characteristics, without a careful distinction between related yet distinct concepts, \textit{opinion} and \textit{stance}. \textit{Opinion} refers to individual's subjective belief or attitude about a topic or entity. It often reflects a speaker’s evaluation, which may be explicit or implied in language~\citep{Oskamp2005-li}. While \textit{Stance}, in contrast, refers to the speaker’s expressed position or orientation \emph{toward} a specific proposition or opinion. Stance is often shown through agreement, disagreement, or neutrality in response to another utterance~\citep{Bois2007-ps,ALDayel2021-fj}. Therefore, in conversation, a stance is observable alignment that may reflect an opinion, but it can also be situational. In this way, opinions can inform stances, but they remain latent unless made explicit through discourse.

To this end, we evaluate how the implicit opinion affects the follow-up stance in this work. We present a framework to assess the impact of \textit{implicit opinion} in discourse. We examine how stance and certainty cues manifest differently in implicit versus explicit opinionated conversations to uncover subtle patterns of opinion expression. First, we establish the framework to validate \textit{normative alignment}, in which a unified expectation guides appropriate responses for equitable inclusion. This expectation stems from normative discourse principles~\citep{habermas2015theory,grice1975logic}, where toxic language (e.g., hate, dehumanization, or extreme ideological views) is not treated neutrally but is instead met with opposition. By aligning stance judgments with this expectation, we can measure whether models reinforce or resist harmful views, especially when they are expressed implicitly. Then, we highlight key turning points in multi-turn dialogues where stance certainty changes, providing insights into how opinions evolve throughout the conversation. Finally, we show that incorporating implicit turns into computational models affects stance classification performance, illustrating how such inclusion can either amplify or mitigate the expression and identification of opinions. 

\section{Related work}
\textbf{Opinion and Bias Representation}. Implicit opinion bias has been defined as the use of subtle language, including hedging, implicature, and abstraction, which can preserve or amplify social stereotypes even in the absence of explicit prejudice~\citep{Maass1999-ay,Tannen1993-nf}. Most previous work on opinion and bias has focused on direct, explicit social biases, such as gender disparities in word embeddings~\citep{Cheng2022-tn} or demographic biases in LLMs~\citep{Hedderich2025-se}. Several studies have also examined the racial aspect of bias~\citep{Hofmann2024-ml,Sun2025-vu}, often operationalized through identity-linked prompts or response disparities on tone or sentiment polarity. For instance, the study by~\citep{jung2024fairness-aware-5eb} developed fairness-aware methods for online Positive-Unlabeled (PU) learning to address bias and ensure equitable outcomes in machine learning models trained on partially labeled data. Additionally, the study by~\citep{Hedderich2025-se} employed a human-centered framework, focusing on explicit linguistic cues and extracting token-level patterns that highlight systematic shifts, such as the use of gendered pronouns.

More recently, there has been a shift towards addressing the implicit biases, which are not overtly expressed but encoded through subtle cues. Studies such as~\citep{Wen2025-wc,borah2024towards-ea8,Kumar2024-ru,aldayel-etal-2024-covert,Tan2025-dp} analyze the presence of implicit biases in single-turn conversations, revealing that LLMs frequently fail to flag or respond adequately to covertly prejudiced language. Another study by~\citep{rescala2024can-ecd} used the 2019 argument dataset to examine the LLMs' responses (single-turn) and their convincing attributes. A recent study by~\citep{lake2025from-f40} analyzed the post-alignment distributional shift of LLM responses using open-ended QA datasets. The study finds that alignment reduces surface-level diversity while increasing the comprehensiveness of single responses. Thus, they define the stance as the response confirmation of the question-answer as ``both'', ``yes'', or ``no''. Arora et al.~\cite{Hofmann2024-ml} frame the implicit racial bias in LLMs by prompting models with identity-linked names and contexts, revealing disparities in sentiment and response quality across demographic groups. The study by~\citep{Tan2025-xb} explores model alignment through the analysis of implicit preferences as latent social values, which are inferred from community engagement patterns found in user-generated content. \cite{ryan2024unintended-6e2} examined the effect of aligning language models to specific preference sets and shows that the alignment of language models is not a One-Size-Fits-All. Multi-turn conversational stance dynamics have also been explored, as seen in~\cite{fleknoyearusdc-236}, where ``dogmatism'' is assessed through evolving stances. More precisely, the study tracks how users shift their stances across Reddit conversations and classifies their overall dogmatism based on these evolving stances.

\textbf{Framing Implicit Opinion Through Subtle Language}. Upon examining the effect of implicit language, prior work has explored how subtle cues influence the interpretation of tasks, such as the interpretation of superlative comparisons ~\citep{pyatkin-etal-2025-superlatives}. A notable line of research investigates the general effects of linguistic subtlety, such as the use of superlatives or indirect references ~\citep{pyatkin-etal-2025-superlatives,Liu2023-ws}. 
Another work extensively studied the identification of implicit hatespeech~\citep{Sap2020-bp,ElSherief2021-br} or Sarcasm detection in dialogue using subtle cues~\citep{Ghosh2017-hh}. In these studies, implicitness is often assessed based on the surface representation, on whether the target group is explicitly mentioned.  In opinion-focused tasks, recent work~\citep{liebeskind2024navigating-937} explores how LLMs distinguish between explicit and implicit opinions, revealing limitations in current detection strategies and proposing prompt-based improvements. The study by ~\citep{liebeskind2024navigating-937} analyzes the ability of LLMs to generate and distinguish between explicit and implicit opinions, highlighting limitations in identifying implicit opinion content and proposing prompt-based strategies for improvement.

\paragraph{Implicit Stance and Response Dynamics.} 
A complementary line of research focuses on detecting implicit stance, focusing on identifying the speaker’s subtly expressed position as implicit stance, specifically as an indirect reference to targets. For example, \citet{Liu2023-ws} extends the stance triangle framework to incorporate implicit and explicit target relationships, enriching stance data annotations to improve out-of-domain generalization. Additionally, the work by~\citep{Gatto2023-dt} proposed text encoders that leverage Chain-of-Thought prompting and evaluate the performance of ChatGPT and Llama2 in identifying stance using the Semeval2016 dataset. Another framing used a single categorization of bias, "Gender bias," such as the work in \cite{Zhao2024-mv}, which investigates gender bias in LLMs using self-reflection prompts. The study shows that models are more accurate in recognizing bias when gender is explicitly mentioned than when it is implied through indirect cues. 

In contrast to prior work, we present a detailed examination of implicit opinion in various conversational settings. Furthermore, we distinguish our work by grounding the treatment of such subtle cues in a \textit{normative alignment} framework. Rather than treating implicit content as ambiguous or neutral, we assess whether the stance of the responses upholds socially expected norms (e.g., disagreement with extreme or harmful views). 
\section{Experimental Setup}
To examine the concept of opinion inclusion, we evaluate two types of conversational alignments: 1) Surface Explicit Alignment, and 2) Latent Underlying Alignment, where latent implicit opinions are included. We represent a framework relying on \textbf{Normative Alignment}, in which the expectation is that conversational models and human participants respond to content in ways that uphold socially acceptable norms~\citep{habermas2015theory,grice1975logic}. In the context of this study, we define normative alignment as the consistent rejection of toxic or harmful viewpoints. This setting defines implicit conversations based on the severity of the targeted opinion, categorizing them as implicitly toxic, explicitly toxic, or neutral. This categorization helps establish a consistent expectation regarding the appropriate stance toward each type of conversation. Typically, the expected stance toward implicit or explicit toxic content is disagreement, whereas neutral content may warrant more relaxed stances, such as agreement or neutrality. By adopting a normative agreement lens rather than treating human disagreement as noise, we view it as a meaningful signal of a normative stance that is often missing in LLM outputs.

Importantly, these definitions are adapted to reflect the structural and rhetorical complexity found in two distinct conversation environments: (a) LLM chat-based and (b)  human dialogues. In LLM chat settings, implicit toxicity often manifests through indirect instruction, e.g., ``write me a story'' or ``tell me a joke'', that conceal the target within a creative or instructive frame. Conversely, in human dialogues, implicit language tends to emerge through more nuanced comparisons, rhetorical framing, or coded expressions, rather than directly or indirectly stating the target of an opinion
~\citep{Tannen1993-nf}. To account for this, we extend our definition of implicit language to include instances where the target is referenced, but the conclusion is conveyed subtly, without overt expression (Appendix~\ref{app:taskformation} and~\ref{append:annotation_process} explain the annotation guideline). 
\begin{table}[h!]
\centering
\small
\renewcommand{\arraystretch}{1.2}
\begin{tabular}{lcc}
\toprule
\textbf{Source} & \textbf{Turns} & \textbf{Unique Pair Conv.} \\
\midrule
Human (Expert)        & 4210                & 2105 \\
Human   & 1896                & 948  \\
LLM                  & 1140                & 570  \\
\midrule
\textbf{Overall}     & \textbf{7246}       & \textbf{3623} \\
\bottomrule
\end{tabular}
\caption{Overview of the dataset sources and dialogue set. Each pair conversation refers to a user-assistant exchange.}
\label{tab:dataset-overview}
\end{table}

\subsection{Data Collection}
To evaluate the implicit opinion in a conversation set, for human conversations, we used~\cite[DialogConan,][]{bonaldi2022human-machine-cf8}, which contains expert human assistants and~\cite[ContextCounter, ][]{albanyan2023not-b79}, which contains open human conversations collected from \textit{X} posts comprising interactions among many users. For LLM-based assistant conversations, we used two benchmark sources of real user queries from an open-source chatbot~\cite[WildChat,][]{Zhao2024-wz} an open-source log of user–LLM interactions and~\cite[ToxicChat,][]{Lin2023-zt} which focuses on model behavior in toxic conversational contexts. As shown in Table~\ref{tab:dataset-overview}, the total number of turns is approximately 7K across all sources, with the conversations ranging from 2 to 7 turns per exchange.
These datasets provide a solid dialogical data baseline and support our experiment's aim to investigate the interaction type and context of replies. Then, we used LabelBox to initiate two tasks: labeling the Assistant and User stance, along with implicit extreme opinion (implicitly or explicitly toxic opinion). Annotators were trained on pilot examples using Labelbox interface, with consensus labeling of stance (Agree/Disagree/Neutral), opinion type (Implicit vs Explicit Toxicity), and Certainty (Certain, Uncertain, Refuse to Engage, None). The full annotation guidelines and anonymized examples are described in Appendix~\ref{append:annotation_process}, and the corresponding dataset is publicly available on OSF.\footnote{\url{https://osf.io/2azn5/files?view_only=bd8f756bb2e849a1b5102953cf33a775}}.
\begin{figure*}[h!]
    \centering
    \begin{subfigure}{0.80\textwidth}
        \centering
        \includegraphics[width=\linewidth]{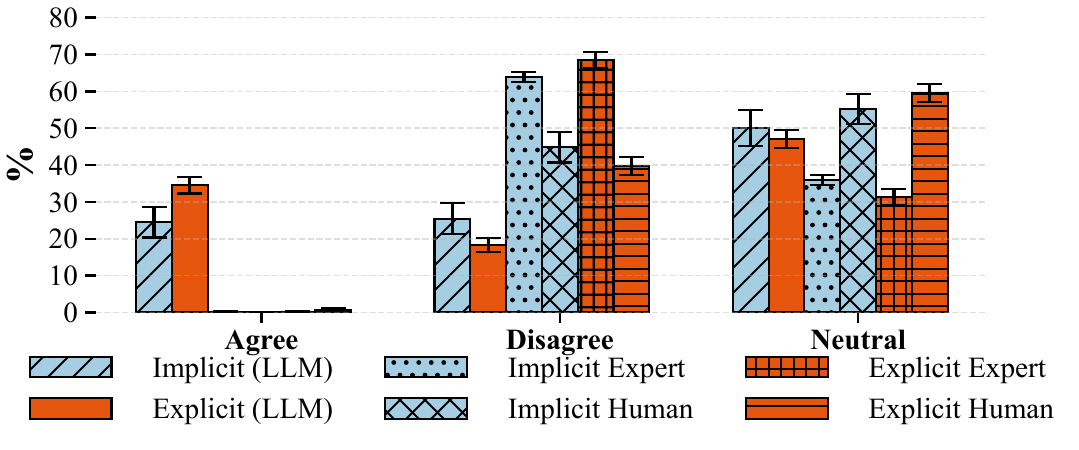}
        \caption{Stance for implicit and explicit opinion turns}
    \end{subfigure}
    \caption{The assistant stance responses (Agree, Disagree, Neutral) across different user input types and sources. The figure compares responses to implicit and explicit prompts from LLMs, expert humans' responses (Expert), and non-expert humans (human). All comparisons show statistically significant results using the chi-square test  $p < .001$}
    \label{fig:stance_group1}
\end{figure*}

\begin{figure}[h!]
    \centering
    \includegraphics[width=1\columnwidth,keepaspectratio]{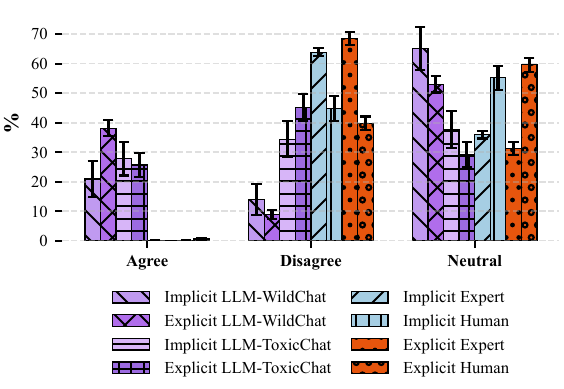}
    \caption{The assistant stance responses compare responses to implicit and explicit per LLMs (WildChat using ChatGPT) and (ToxicChat using Vicuna-7B). 
    All comparisons show statistically significant results using the chi-square test ($p < .001$).}
    \label{fig:stance_group2}
\end{figure}

\begin{figure*}[h]
  \centering
  \begin{subfigure}[t]{0.49\textwidth}
    \centering
    \includegraphics[width=\linewidth]{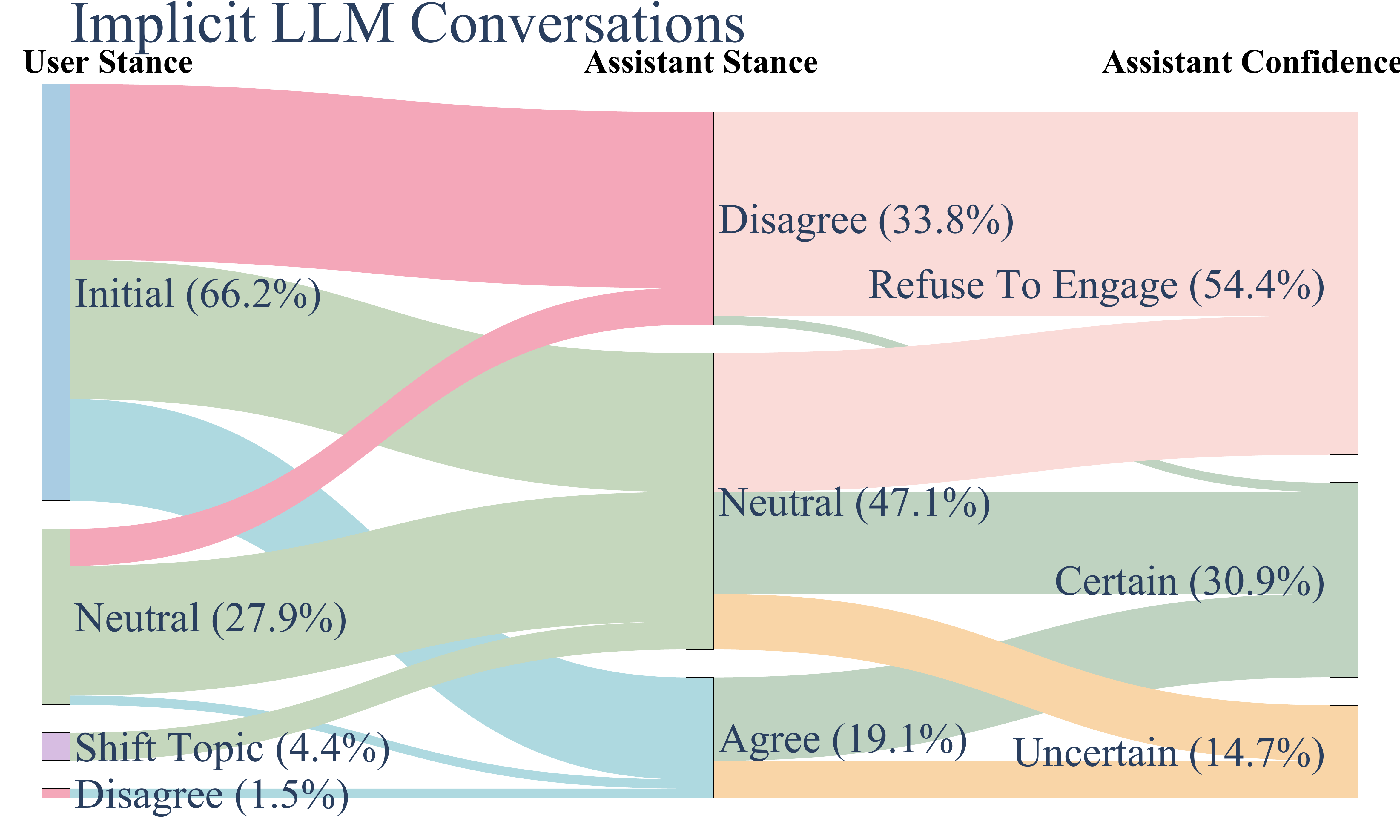}
    \caption{LLM Implicit Conversations}
    \label{fig:sankey_llm_implicit}
  \end{subfigure}
  \hfill
  \begin{subfigure}[t]{0.49\textwidth}
    \centering
    \includegraphics[width=\linewidth]{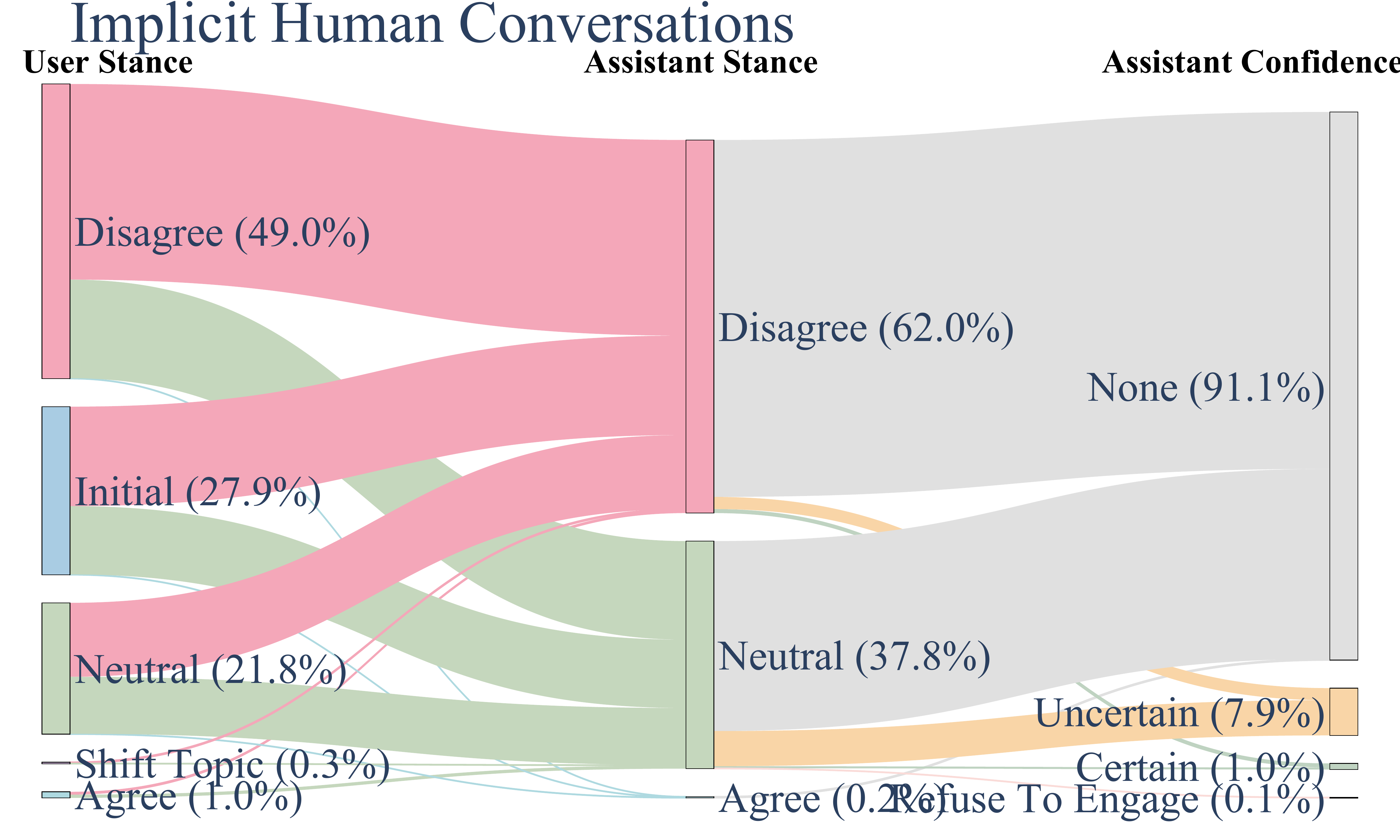}
    \caption{Human Implicit Conversations}
    \label{fig:sankey_human_implicit}
  \end{subfigure}
  \caption{Stance transitions in implicit conversations where each flow begins with the user’s stance, moves through the assistant’s stance, and ends with the assistant's confidence. Certainty labels capture the assistant’s expressed confidence: Certain, Uncertain, Refuse to Engage, and None. \% are the relative distributions at each node.}
  \label{fig:sankey_combined}
\end{figure*}
\subsection{Inclusive Implicit Learning models}
We evaluate two learning paradigms to assess the model’s ability to internalize subtle opinion cues: 1) post-training using Instruct Tune on implicit conversations using decoder-only LLMs and 2) positive-unlabeled (PU) online learning using linear and shallow neural models trained on Sentence-BERT embeddings. In both setups, the training data includes varying proportions of implicit opinion examples, ranging from 10\% to 100\%, to evaluate scalability and robustness. Zero-shot and 0\% implicit training settings are included as lower baselines. As the implicit opinions usually remain unlabeled or are harder to annotate. This case of scarcity of unlabeled examples has been extensively studied as a Positive-Unlabeled (PU) learning scenario~\citep{jung2024fairness-aware-5eb}, with a focus on explicitly mentioning the target group. Instead, our study examines another angle of implicit and subtle reference to opinion. Thus, we formulate positive samples to include explicitly labeled stances, while unlabeled samples include texts with potential implicit stances (which might be Agree or Disagree). We formulate our task as a binary stance classification problem between \texttt{Agree} (positive class) and \texttt{Disagree} (negative class). Only these two stance categories are retained during preprocessing. In (PU) training, for each assistant response, we concatenate the user and assistant messages (user [SEP] assistant) and represent them using dense semantic embeddings from a Sentence-BERT model (all-MiniLM-L6-v2). As for LLMs (Llama3 and Mistral), we used an instruction tuning prompt that includes the context of the user's implicit opinion. For reproducibility, we used Mistral-7B-Instruct-v0.3 and LLaMA-3-8B-Instruct models, each fine-tuned with LoRA adapters on implicit conversation data. Complete hyperparameters and model parameter sizes are summarized in Appendix~\ref{app:models_expSetup} (Tables 7–8). 
\paragraph{Implicit Group Sensitive PU-style setup.}
We adopt principles from positive-unlabeled learning~\citep{jung2024fairness-aware-5eb} to handle imbalance and fairness settings between implicit and explicit contextual opinion expressions. Each example is tagged with a sensitive attribute based on whether the user message expresses an implicit (represented as 0) or explicit (represented as 1) opinion. These group indicators are used to handle fairness constraints in PU, ensuring that models maintain comparable false positive rates (FPR) across both implicit and explicit opinions.

\section{Results}
We begin by presenting the results of analyzing the interplay between stance in various implicit and explicit conversations between humans and LLM assistants \S\ref{subsec:user_behavior}. Then, in section \S\ref{subsec:model_behavior}, we detail the result of our portion of implicit training. 
\subsection{Evaluating Normative Alignment in Implicatures Conversations}
\label{subsec:user_behavior}
First, we evaluate how well conversational responses align with social norms when implicatures are used in real conversations to convey the meaning indirectly or implicitly, rather than explicitly.
To do so, we analyze the real stance of the responses across discourse (LLM-generated and human assistant responses) using that as a means to evaluate the norm alignment. Referring to our experiment design, we used the extreme cases of the harmful implicit/explicit cases to unify the expected behavior of LLMs and human assistant responses. 
\paragraph{ Assistant stance in response to implicit opinion.} We demonstrate the interplay between human and LLM responses in various scenarios to compare the distinct behavior of assistance stance between implicit and explicit opinion as shown in Figure~\ref{fig:stance_group1}. In general, humans tend to follow the normative expectation of disagreeing with toxic content, especially when discourse is explicit. In particular, Expert humans show high disagreement rates toward explicit opinion, reflecting a stronger normative alignment. Interestingly, LLMs have a higher likelihood of agreement when the conversation explicitly has a harmful opinion, potentially due to surface-level alignment. In contrast, responses to implicit discourse elicit more neutral stances from LLMs, suggesting hesitation or ambiguity in detecting subtler expressions. All comparisons are statistically significant based on a chi-square test ($p < .001$) as shown in Appendix~\ref{app:chi_comparision}. To verify that these patterns are not artifacts of mixed datasets, we analyzed the two LLM subsets separately (Figure~\ref{fig:stance_group2}).Across both WildChat (ChatGPT) and ToxicChat (Vicuna-7B), Figure~\ref{fig:stance_group2} replicates the pattern in Figure~\ref{fig:stance_group1}. Particularly, LLMs show higher agreement with explicit harmful opinions and more neutral responses to implicit ones, confirming surface-level rather than contextual norm alignment. Across models, ChatGPT (WildChat) exhibits slightly higher neutrality, whereas Vicuna-7B (ToxicChat) shows stronger disagreement with explicit opinions. Moreover, we analyze the certainty markers associated with the assistance responses. As shown in figure~\ref{fig:sankey_combined}, LLMs tend to respond more cautiously, using “Refuse to Engage” or neutral tones more often, and expressing certainty more explicitly. In contrast, humans disagree openly in implicit contexts but rarely tag their certainty. Overall, the vast majority of human certainty is marked as None (91.1\%), indicating that humans do not explicitly express certainty as often. 
\paragraph{Flow of the stance and certainty markers}
As shown in Figure~\ref{fig:sankey_combined}, LLMs tend to respond more cautiously, using “Refuse to Engage” or neutral tones more often, and expressing certainty more explicitly. In contrast, humans disagree openly in implicit contexts but rarely tag their certainty. Overall, the vast majority of human certainty is marked as None (91.1\%), indicating that humans do not explicitly express certainty as often. 
\begin{figure}[b!]
    \centering
    \begin{subfigure}[t]{0.90\linewidth}
        \centering
        \includegraphics[width=\linewidth]{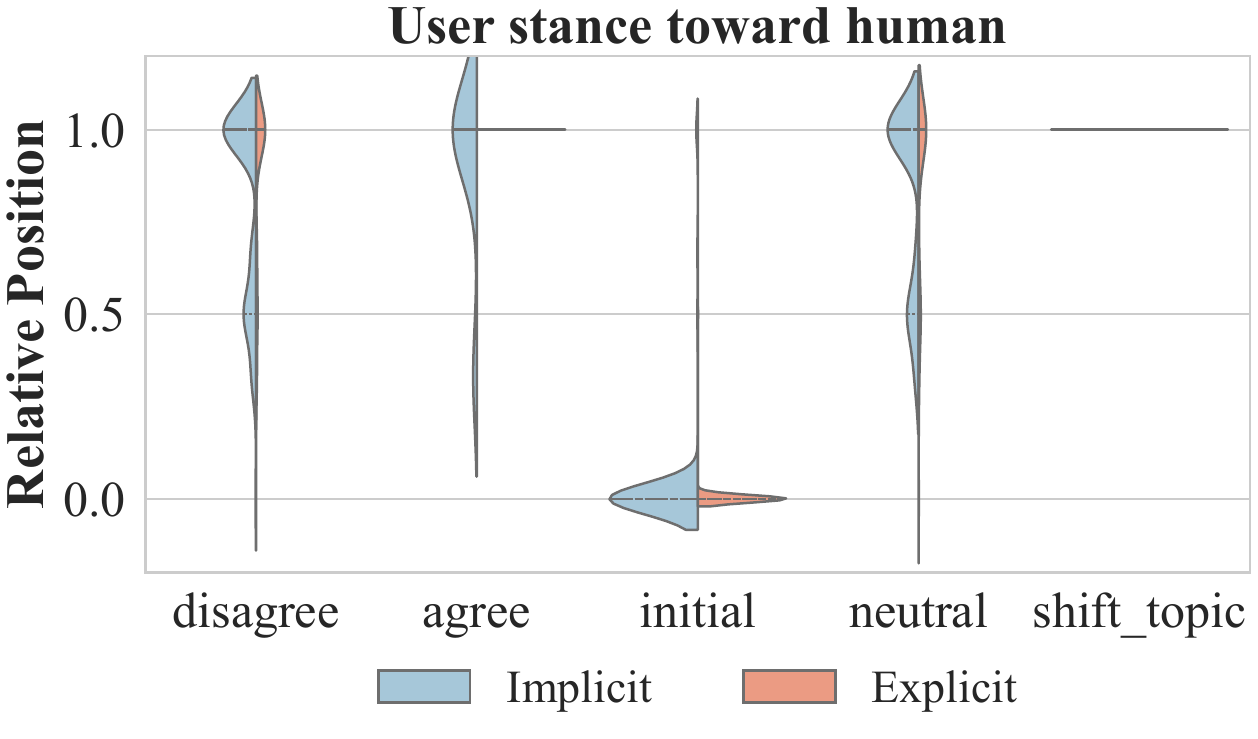}
        \caption{User stance towards human responses}
        \label{fig:stance_violin_human}
    \end{subfigure}
    \begin{subfigure}[t]{0.90\linewidth}
        \centering
        \includegraphics[width=\linewidth]{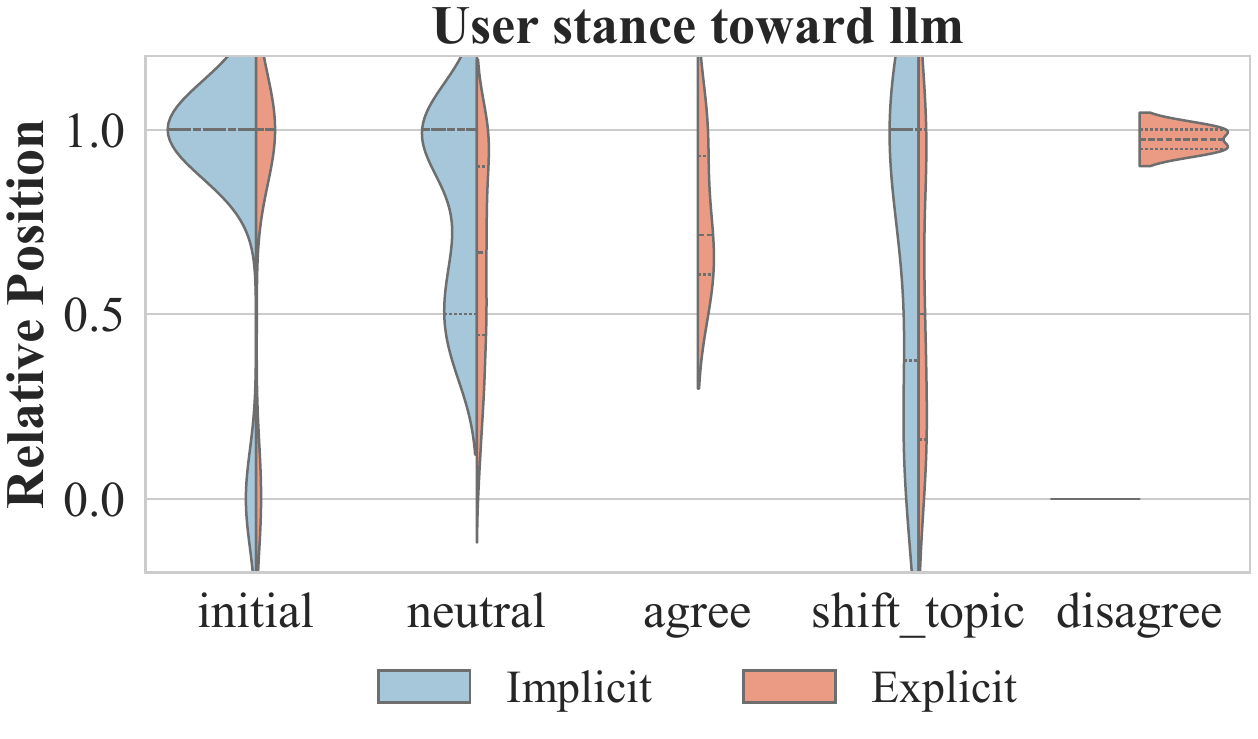}
        \caption{User stance towards LLM responses}
        \label{fig:stance_violin_llm}
    \end{subfigure}
    \caption{
        The relative position of user stances across conversations with human and LLM-generated responses. The y-axis is normalized position of each turn within the conversation (\texttt{turnID$/$TotalLength}), where 0.5 marks the midpoint.  
    }
    \label{fig:stance_violin_combined}
  
\end{figure}

\begin{table*}[ht!]
\centering
\resizebox{\textwidth}{!}{%
\begin{tabular}{llccccccc}
\toprule
\multirow{2}{*}{\textbf{Method}} & \multirow{2}{*}{\textbf{Model}} & \textbf{Zero-Shot} & \textbf{0\%} & \textbf{10\%} & \textbf{20\%} & \textbf{30\%} & \textbf{60\%} & \textbf{100\%} \\
\cmidrule(lr){3-9}
& & \multicolumn{7}{c}{\textbf{Macro F1 Score ± Std}} \\
\midrule
\multirow{2}{*}{Fine-tuning} 
  & LLaMA3  & 0.462 $\pm$ 0.026 
           & \cellcolor{red!20} 0.423 $\pm$ 0.0717 
           & 0.434 $\pm$ 0.1345 
           & 0.464 $\pm$ 0.0259 
           & 0.399 $\pm$ 0.1633 
           & \cellcolor{green!20} 0.487 $\pm$ 0.0378 
           & 0.480 $\pm$ 0.0357 \\
  & Mistral & \cellcolor{red!20}0.131 $\pm$ 0.002
           & 0.942 $\pm$ 0.003
           & 0.936 $\pm$ 0.002
           & 0.941 $\pm$ 0.003
           & 0.940 $\pm$ 0.003
           & 0.930 $\pm$ 0.003
           & \cellcolor{green!20}0.944 $\pm$ 0.002 \\
\midrule
\multirow{3}{*}{\shortstack[l]{Positive-Unlabeled}} 
  & Linear & - & 0.764 $\pm$ 0.066  
           & \cellcolor{green!20}0.775 $\pm$ 0.076  
           & 0.737 $\pm$ 0.069  
           & 0.738 $\pm$ 0.068  
           & 0.695 $\pm$ 0.076  
           & \cellcolor{red!20}0.654 $\pm$ 0.107  \\
  & Mlp    & - & 0.202 $\pm$ 0.027  
           & \cellcolor{green!20}0.208 $\pm$ 0.039  
           & 0.202 $\pm$ 0.059  
           & 0.182 $\pm$ 0.034  
           & 0.197 $\pm$ 0.026  
           & \cellcolor{red!20}0.191 $\pm$ 0.052  \\
\bottomrule
\end{tabular}
}
\caption{Macro F1 scores across varying percentages of implicit data and averaged over 5 folds. Best (green) and worst (red) scores are highlighted.}
\label{tab:macro_f1_std_all_models}
\end{table*}

\paragraph{Stance transitions in implicit conversations }
We analyze the turning point of stance within the conversation as shown in Figure~\ref{fig:stance_violin_combined}. Mainly, it illustrates the distribution of user stance positions within conversations involving human and LLM-generated responses. The y-axis represents the normalized position of each user's turn, with higher values indicating later turns. Across both assistant types, \textit{agree} and \textit{neutral} stances are expressed in later parts of the conversation. However, two key patterns can be noticed. First, \textit{initial} stances in human dialogues occur significantly earlier when users hold implicit opinions, indicating an early assertion of viewpoint in the face of ambiguity. Second, users are more likely to express \textit{disagree} stances earlier in conversations with humans than with LLMs, especially when opinions are implicit. For LLMs, the only significant shift appears in the \textit{neutral} stance, where users with implicit opinions tend to reach neutrality earlier. These patterns suggest users exhibit greater conversational assertiveness, either through disagreement or early opinion assertion when responding to human assistants, while interactions with LLMs show more delayed or neutral positioning. We validated the significance of our comparison and conducted Mann-Whitney U tests comparing the relative timing of user stances between explicit and implicit opinion contexts (Appendix~\ref{app:chi_comparision}). 

\subsection{Model Performance across Implicit Training Portions}
\label{subsec:model_behavior}
As shown in Table~\ref{tab:macro_f1_std_all_models}, Mistral achieves consistently strong macro F1 scores across all inclusion levels, with performance peaking at 100\% implicit inclusion ($0.944$). In contrast, LLaMA3 lags behind, particularly at lower inclusion levels. As for PU models, the linear classifier performs robustly at low inclusion (10\%: $0.775$), while the MLP shows high variance and degraded performance.

\begin{figure}[h]
    \centering
    \includegraphics[width=0.90\linewidth]{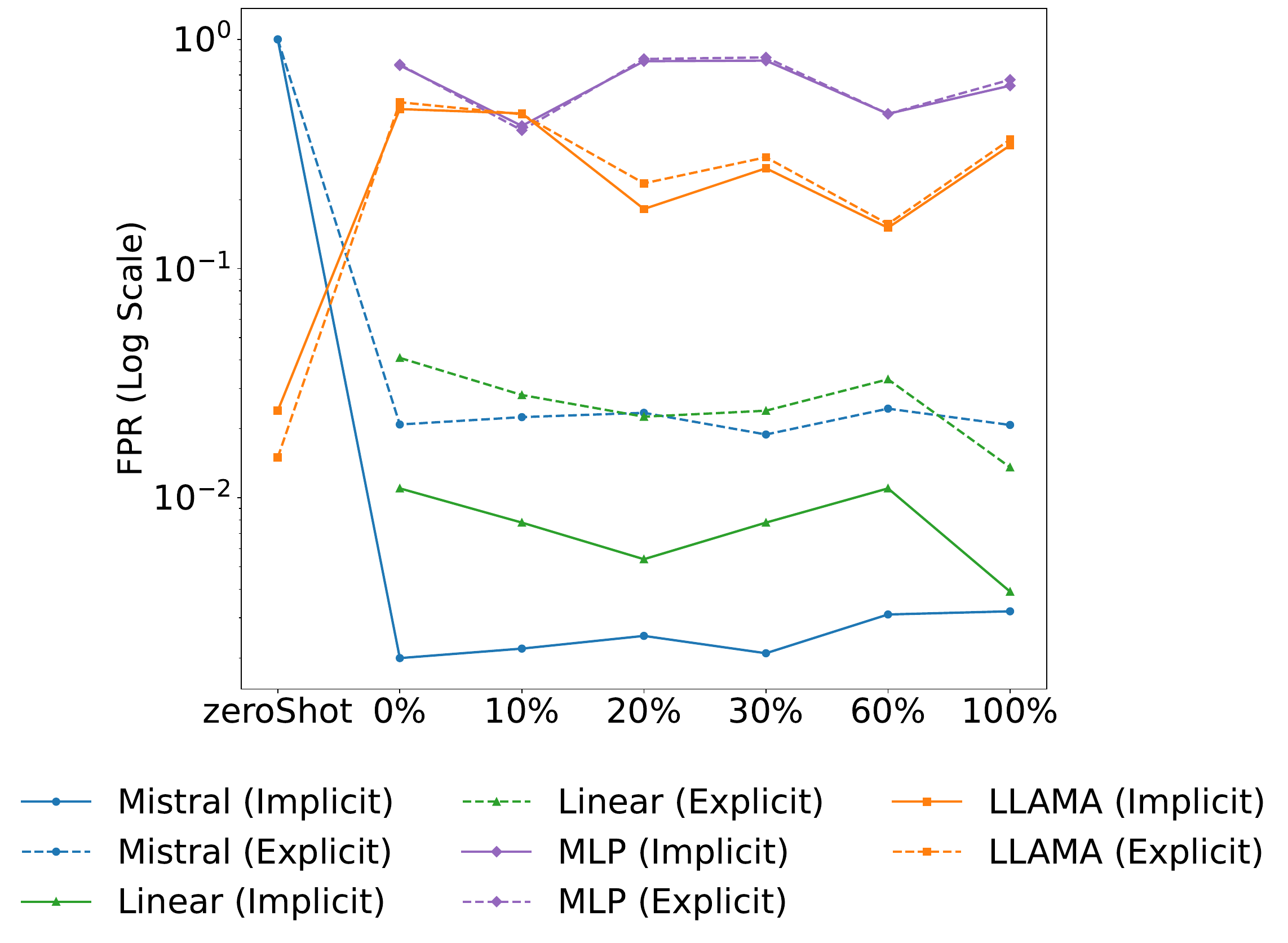}
    \caption{False Positive Rate (FPR) across different portions using a logarithmic scale. }
    \label{fig:fpr_mistral_log_plot}
\end{figure}

Figure~\ref{fig:fpr_mistral_log_plot} complements these results by showing that both Mistral and Linear models maintain low false positive rates (FPR), especially beyond 10\% inclusion. Notably, MLP models exhibit a sharp spike in FPR at 0\% and zero-shot settings, underscoring their inability to generalize without the inclusion of implicit cues. This overprediction of the \texttt{Agree} class in norm-sensitive contexts demonstrates poor calibration and indicates risk of norm violation. In contrast, LLaMA3 maintains a low FPR at these early settings, but this is linked with low macro F1 scores (see Table~\ref{tab:macro_f1_std_all_models}), suggesting underprediction or overly conservative behavior rather than calibrated learning, which is a different type of failure mode. When comparing the implicit and explicit False Positive Rates (FPRs) within each model, both PU-learned models (Linear and MLP) and LLaMA3 exhibit a relatively small FPR gap across different discourse styles. This indicates that these models behave consistently, regardless of whether the language used is subtle or overt. In contrast, the Mistral model exhibits a larger FPR gap, especially at low inclusion levels, which suggests a bias toward surface-level (explicit) cues. 
The narrower False Positive Rate gap observed in PU models and LLaMA-3 indicates more balanced treatment of implicit and explicit language, suggesting higher fairness consistency. This low FPR is due to different reasons, as PU models benefit from fairness-aware training of implicit and explicit groups. While LLaMA3’s uniform behavior of FPR between explicit and implicit opinions suggests that the model tends to adopt a conservative stance by avoiding agreement even when it may be the correct stance. 

\section{Discussion and Implications}
\paragraph{Role of implicatures in communication.} Our findings reveal that LLMs exhibit surface-level alignment by agreeing with explicitly toxic opinions. At the same time, humans maintain stable normative disagreement. This indicates that model alignment remains superficial when norms are implied rather than overtly stated. As illustrated in Figures~\ref{fig:stance_group1} and~\ref{fig:sankey_combined}, assistant stance response behaviors differ across implicit and explicit user opinions. Figure~\ref{fig:stance_group1} reveals that LLMs have a higher rate of agreement with explicit extreme toxic opinion, compared to implicit toxicity. While, expert assistants humans show more stable disagreement regardless of implicit or explicit misaligned norms. By zooming in on the neutral user opinion as shown in Figure~\ref{fig:stance_neutral_combined}, Appendix~\ref{app:taskformation}, human assistants are more likely to disagree, while LLMs tend to still be agreeable. This confirm our experiment design to focus on extreme clear cases of implicit toxic opinion to facilitate the overall examination of stance as a means to evaluate the social norms. This behavior of complicity in LLMs' responses, even toxic opinions, has been underscored in previous studies as ``sycophancy''~\citep{Hong2025-pz,Cheng2025-wk,Rrv2024-bq}, where LLMs show agreeable behavior with users' statements. However, our findings extend this line of work by examining agreement in the presence of implicit opinion cues, such as implicatures and indirect expressions of norm misaligned context (toxicity). Unlike prior studies that focus on overt stance shifts, we demonstrate that LLMs remain agreeable even in subtly toxic or norm mismatch contexts, particularly when opinions are implicitly framed. 
Complementing this, Figure~\ref{fig:sankey_combined} shows that in implicit opinions, human assistants tend to respond to initial or neutral stances with clear disagreement and high certainty, whereas LLMs often either refuse to engage or express uncertainty. These trends underscore a normative alignment gap in LLM responses, where human assistants tend to maintain a more decisive and oppositional stance toward problematic content that is toxic. At the same time, LLMs display an inconsistent stance of neutrality and usually tend to agree with those toxic misaligned norms and signaling, and lower certainty when facing implicit toxicity. In contrast to prior work that analyzes certainty markers in isolation~\citep{Rottger2024-vv}, our analysis reveals that focusing solely on refusal or uncertainty overlooks how models may simultaneously express stance alignment, espicially the neutral or agreeable stances toward norm violating content. This behaviour sheds light on a fixed or superficial reliance on neutral responses, which might not be a sufficient safeguard in value sensitive conversations, especially when toxicity or bias is embedded through implicature or indirect opinion expression. Our findings advocate for integrating stance analysis with certainty calibration to better evaluate normative alignment in implicit contexts.

\begin{figure}[th!]
  \centering

  \begin{subfigure}{0.23\textwidth}
    \centering
    \includegraphics[width=\linewidth]{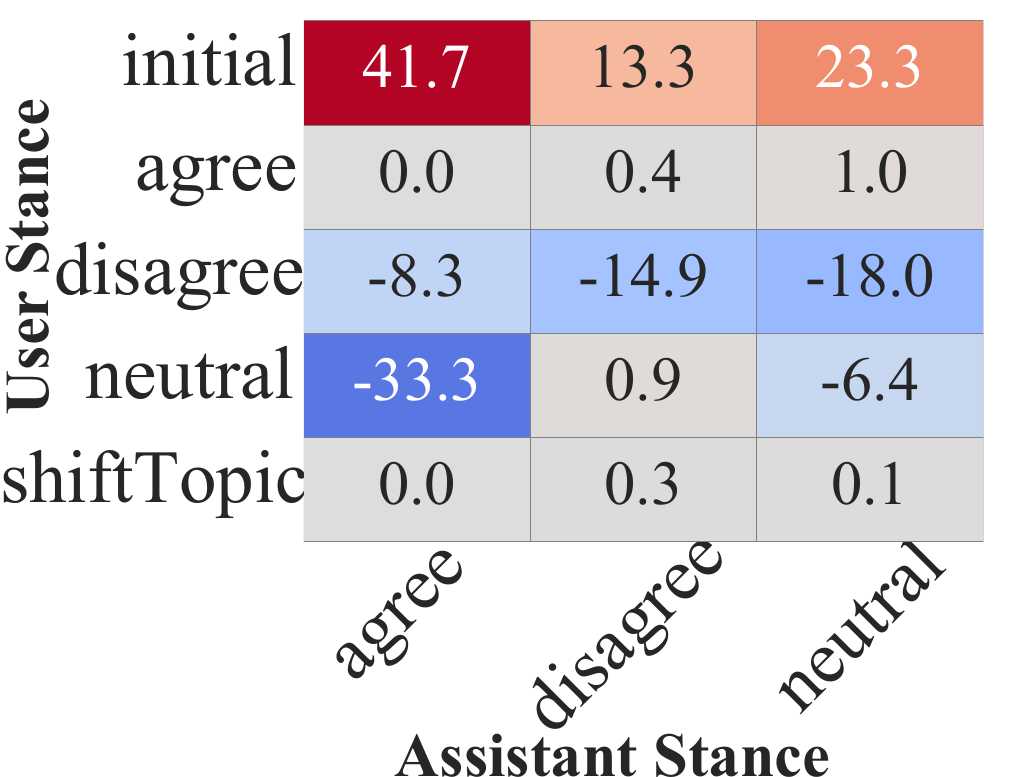}
    \caption{Human responses}
  \end{subfigure}
  \begin{subfigure}{0.20\textwidth}
    \centering
    \includegraphics[width=\linewidth]{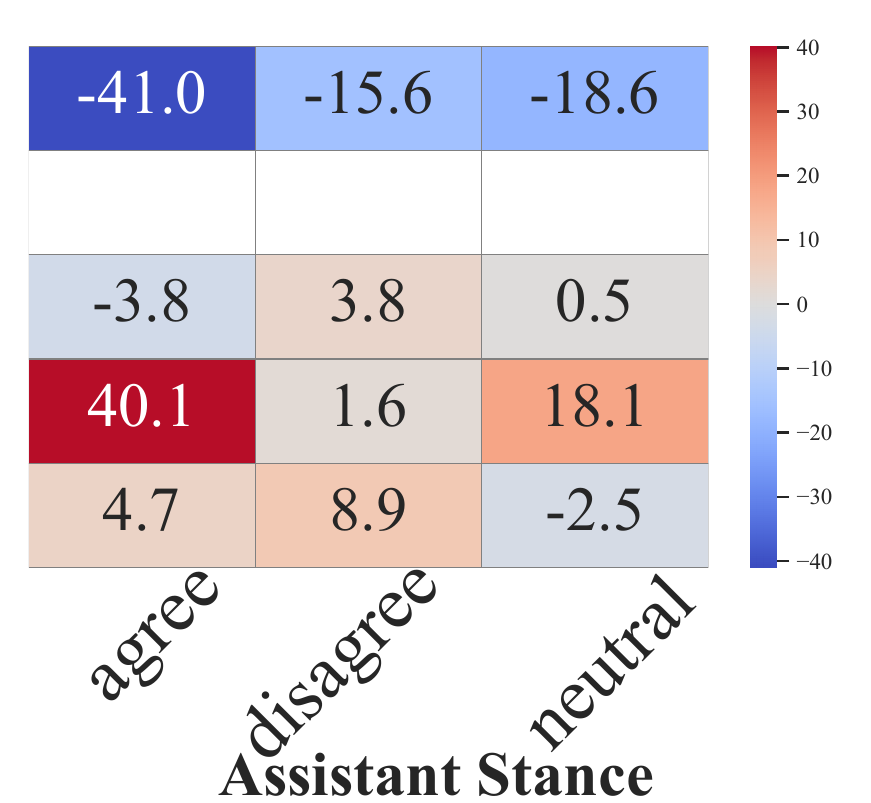}
    \caption{LLM responses}
  \end{subfigure}
  \caption{Difference \% in user stance distributions (given assistant stance) between explicit and implicit opinion. Positive values indicate higher proportions in the explicit condition.}
  \label{fig:dif_exp-vs-imp}
\end{figure}

\textbf{Dynamics of user stance within conversation narratives} Building on our examination of user stance toward assistant replies, we further analyze the user reaction towards the assistant. As shown in figure~\ref{fig:dif_exp-vs-imp}, it can be seen that when extreme opinion is overt, users recognize the assistant’s corrective or balanced stance and respond supportively. On the other hand, implicit extreme opinion has greater user disagreement, potentially because the harm is debatable (Especially toward human assistants). Unlike human assistants, LLMs provoke more user disagreement when responding to explicit extreme opinions, particularly when they remain neutral. This behavior can be explained through the ``Elaboration Likelihood Model (ELM) of persuasion''~\citep{Petty1986-ub}, which states that attitude change occurs through either central (deep) or peripheral (surface-level) processing route. In the case of LLM generated responses, users may fail to engage in deep processing if the assistant's message lacks perceived credibility or personal relevance. Instead, users usually rely on peripheral cues (surface-level), such as tone of certainty or refusal to engage in the conversation, as we observed in Figure~\ref{fig:sankey_combined}, where the LLM frequently adopts a refusal tone. Consequently, users are less likely to shift their stance or revise their disagreement in response to the LLM, unless it presents highly credible or reasoned arguments. This tendency has been supported in various ways by surface-level stance interactions~\citep{Aldayel2022-ew} or as demonstrated by ~\citep{Gallegos2025-zz} through a user's perspective on labeled AI responses.

\paragraph{The magnitude scalability of implicit training.} 
A closer examination of the results in Table~\ref{tab:macro_f1_std_all_models} and Figure~\ref{fig:fpr_mistral_log_plot} shows that scaling the inclusion of implicit conversational data results in measurable improvements in both performance and calibration of FPRs. As can be seen, Mistral's overall performance is enhanced compared to zero-shot and remains robust at partial training levels, achieving high F1 scores (above 0.93 from as low as 10\% inclusion) while maintaining a low False Positive Rate (FPR), which is an indicator of reliability in norm sensitive classification. Additionally, linear PU models trained on implicit opinions consistently exhibit low FPRs, demonstrating the benefits of even shallow architectures (linear) when trained on implicit supervision. In contrast, deeper MLP models remain less reliable, with high FPRs, suggesting that more complex models may require additional regularization or architectural adjustments to handle implicit nuance effectively (topic-level discrimination). The 30\% inclusion portion is a critical threshold as below that threshold, models struggle with implicit opinion patterns. While above 30\%, Mistral and Linear models show consistent model behavior. We further verify the performance between models comparisons using MacNamer's test in Tables~\ref{tab:portionsmcnemar-comparisonllam3_mistral},~\ref{tab:portionsPUmcnemar_avg} (Appendix~\ref{Apd:McNemarbetweenmodels}) which support this behaviour with evidence of reduced overgeneralization errors past this point. Thus, a monitored and balanced inclusion of implicit data improves accuracy and minimizes false agreement with harmful perspectives.

\section{Conclusion}
Achieving equitable inclusion that aligns with normative standards requires addressing implicit expressions of opinion. This study empirically evaluates opinion exchange within realistic conversational turns and considers its impact on the follow-up stance. These findings underscore the importance of incorporating implicit conversations into training and conversational norm based evaluations. Rather than treating them as exceptions, their inclusion helps create socially aware models that can recognize subtle cues and maintain value-sensitive behavior in diverse communication contexts. 

\section*{Limitations}
The datasets used such as DialogConan, ToxicChat may reflect sociocultural norms that are specific to certain communities or platforms. Thus, the generalizability of the normative alignment framework across diverse cultural and linguistic contexts remains limited and needs to be considered in future cross-cultural studies.
Additionally, our social norm evaluation used a few LLMs (Mistral, LLaMA3) transformer-based models that tend to memorize factual and normative patterns from their extensive pretraining.  While human and LLM dialogues originate from distinct contexts, this design intentionally preserves ecological validity, reflecting how implicit opinions naturally arise in real-world conversations. Future work will extend EMBRACE to controlled settings using identical prompts for both humans and LLMs to strengthen cross-domain comparability. However, our current analysis scope does not empirically assess how temporal aspects of model training, or the evolving nature of norms within training data, might impact their alignment with socially expected stances. As a future direction, it would be beneficial to consider examining this aspect, particularly the temporal shifts in normative behavior and their impact on stance consistency.
\section*{Ethics Statement}
 This study aims to advance equitable inclusive of opinion representation in conversational models by including implicit opinion and using stance as means to evaluate normative alignment. Motivated by ethical computing principles such as ACM Code of Ethics Principle 1.4 (“Be fair and take action not to discriminate”), this study seeks to evaluate the implicit language through which conversational models may reinforce norm-violating or harmful views. Although we recognize that any biased or poorly designed community language models can unintentionally reinforce stereotypes. We emphasize that our framework does not view human disagreement as mere noise, but rather as an important reflection of social norms. Our objective is to encourage value-sensitive, inclusive design without silencing diverse yet respectful viewpoints. Annotators were briefed on handling toxic material and provided with opt-out options.

\bibliography{paperpileImp,references,anthology,Hatepaperpile}

@INPROCEEDINGS{Sap2020-bp,
  title     = "Social Bias Frames: Reasoning about Social and Power Implications
               of Language",
  author    = "Sap, Maarten and Gabriel, Saadia and Qin, Lianhui and Jurafsky,
               Dan and Smith, Noah A and Choi, Yejin",
  editor    = "Jurafsky, Dan and Chai, Joyce and Schluter, Natalie and
               Tetreault, Joel",
  booktitle = "Proceedings of the 58th Annual Meeting of the Association for
               Computational Linguistics",
  publisher = "Association for Computational Linguistics",
  address   = "Online",
  pages     = "5477--5490",
  month     =  jul,
  year      =  2020
}

@INPROCEEDINGS{Ghosh2017-hh,
  title     = "The role of conversation context for sarcasm detection in online
               interactions",
  author    = "Ghosh, Debanjan and Richard Fabbri, Alexander and Muresan,
               Smaranda",
  booktitle = "Proceedings of the 18th Annual SIGdial Meeting on Discourse and
               Dialogue",
  publisher = "Association for Computational Linguistics",
  address   = "Stroudsburg, PA, USA",
  pages     = "186--196",
  year      =  2017
}

@INPROCEEDINGS{ElSherief2021-br,
  title     = "Latent Hatred: A Benchmark for Understanding Implicit Hate Speech",
  author    = "ElSherief, Mai and Ziems, Caleb and Muchlinski, David and
               Anupindi, Vaishnavi and Seybolt, Jordyn and De Choudhury, Munmun
               and Yang, Diyi",
  editor    = "Moens, Marie-Francine and Huang, Xuanjing and Specia, Lucia and
               Yih, Scott Wen-Tau",
  booktitle = "Proceedings of the 2021 Conference on Empirical Methods in
               Natural Language Processing",
  publisher = "Association for Computational Linguistics",
  address   = "Online and Punta Cana, Dominican Republic",
  pages     = "345--363",
  month     =  nov,
  year      =  2021
}

@ARTICLE{ALDayel2021-fj,
  title    = "Stance detection on social media: State of the art and trends",
  author   = "ALDayel, Abeer and Magdy, Walid",
  journal  = "Inf. Process. Manag.",
  volume   =  58,
  number   =  4,
  pages    =  102597,
  month    =  jul,
  year     =  2021
}

@INPROCEEDINGS{Zhao2024-mv,
  title     = "A Comparative Study of Explicit and Implicit Gender Biases in
               Large Language Models via Self-evaluation",
  author    = "Zhao, Yachao and Wang, Bo and Wang, Yan and Zhao, Dongming and
               Jin, Xiaojia and Zhang, Jijun and He, Ruifang and Hou, Yuexian",
  booktitle = "Proceedings of the 2024 Joint International Conference on
               Computational Linguistics, Language Resources and Evaluation
               (LREC-COLING 2024)",
  pages     = "186--198",
  year      =  2024
}

@INPROCEEDINGS{Liu2023-ws,
  title     = "Guiding Computational Stance Detection with Expanded Stance
               Triangle Framework",
  author    = "Liu, Zhengyuan and Yap, Yong Keong and Chieu, Hai Leong and Chen,
               Nancy",
  editor    = "Rogers, Anna and Boyd-Graber, Jordan and Okazaki, Naoaki",
  booktitle = "Proceedings of the 61st Annual Meeting of the Association for
               Computational Linguistics (Volume 1: Long Papers)",
  publisher = "Association for Computational Linguistics",
  address   = "Toronto, Canada",
  pages     = "3987--4001",
  month     =  jul,
  year      =  2023
}

@INPROCEEDINGS{Jung2024-pb,
  title     = "Fairness-aware online positive-unlabeled learning",
  author    = "Jung, Hoin and Wang, Xiaoqian",
  booktitle = "Proceedings of the 2024 Conference on Empirical Methods in
               Natural Language Processing: Industry Track",
  publisher = "Association for Computational Linguistics",
  address   = "Stroudsburg, PA, USA",
  pages     = "170--185",
  month     =  nov,
  year      =  2024
}

@INPROCEEDINGS{Lin2023-zt,
  title     = "{ToxicChat}: Unveiling hidden challenges of toxicity detection in
               real-world user-{AI} conversation",
  author    = "Lin, Zi and Wang, Zihan and Tong, Yongqi and Wang, Yangkun and
               Guo, Yuxin and Wang, Yujia and Shang, Jingbo",
  booktitle = "Findings of the Association for Computational Linguistics: EMNLP
               2023",
  publisher = "Association for Computational Linguistics",
  address   = "Stroudsburg, PA, USA",
  pages     = "4694--4702",
  month     =  dec,
  year      =  2023
}

@INCOLLECTION{Petty1986-ub,
  title     = "The elaboration likelihood model of persuasion",
  author    = "Petty, Richard E and Cacioppo, John T",
  booktitle = "Advances in Experimental Social Psychology",
  publisher = "Elsevier",
  volume    =  19,
  pages     = "123--205",
  series    = "Advances in experimental social psychology",
  month     =  jan,
  year      =  1986
}

@INPROCEEDINGS{Kumar2024-ru,
  title     = "Subtle biases need subtler measures: Dual metrics for evaluating
               representative and affinity bias in large language models",
  author    = "Kumar, Abhishek and Yunusov, Sarfaroz and Emami, Ali",
  booktitle = "Proceedings of the 62nd Annual Meeting of the Association for
               Computational Linguistics (Volume 1: Long Papers)",
  publisher = "Association for Computational Linguistics",
  address   = "Stroudsburg, PA, USA",
  pages     = "375--392",
  year      =  2024
}

@INPROCEEDINGS{Tan2025-dp,
  title     = "Unmasking implicit bias: Evaluating persona-prompted {LLM}
               responses in power-disparate social scenarios",
  author    = "Tan, Bryan Chen Zhengyu and Lee, Roy Ka-Wei",
  booktitle = "Proceedings of the 2025 Conference of the Nations of the Americas
               Chapter of the Association for Computational Linguistics: Human
               Language Technologies (Volume 1: Long Papers)",
  publisher = "Association for Computational Linguistics",
  address   = "Stroudsburg, PA, USA",
  pages     = "1075--1108",
  year      =  2025
}

@INPROCEEDINGS{Cheng2022-tn,
  title     = "Debiasing Word Embeddings with Nonlinear Geometry",
  author    = "Cheng, Lu and Kim, Nayoung and Liu, Huan",
  booktitle = "Proceedings of the 29th International Conference on Computational
               Linguistics",
  pages     = "1286--1298",
  month     =  oct,
  year      =  2022
}

@INPROCEEDINGS{Gatto2023-dt,
  title     = "Chain-of-thought embeddings for stance detection on social media",
  author    = "Gatto, Joseph and Sharif, Omar and Preum, Sarah",
  editor    = "Bouamor, Houda and Pino, Juan and Bali, Kalika",
  booktitle = "Findings of the Association for Computational Linguistics: EMNLP
               2023",
  publisher = "Association for Computational Linguistics",
  address   = "Stroudsburg, PA, USA",
  pages     = "4154--4161",
  year      =  2023
}

@INPROCEEDINGS{Wen2025-wc,
  title     = "Evaluating Implicit Bias in Large Language Models by Attacking
               From a Psychometric Perspective",
  author    = "Wen, Yuchen and Bi, Keping and Chen, Wei and Guo, Jiafeng and
               Cheng, Xueqi",
  booktitle = "Findings of the Association for Computational Linguistics: ACL
               2025",
  publisher = "Association for Computational Linguistics",
  address   = "Taipei, Taiwan",
  year      =  2025
}

@INCOLLECTION{Bois2007-ps,
  title     = "The Stance Triangle",
  author    = "Bois, John W Du",
  editor    = "Englebretson, Robert",
  booktitle = "Stancetaking in Discourse: Subjectivity, Evaluation, Interaction",
  publisher = "John Benjamins Publishing Company",
  address   = "Amsterdam/Philadelphia",
  pages     = "139--182",
  year      =  2007
}

@ARTICLE{Hedderich2025-se,
  title         = "What's the difference? Supporting users in identifying the
                   effects of prompt and model changes through token patterns",
  author        = "Hedderich, Michael A and Wang, Anyi and Zhao, Raoyuan and
                   Eichin, Florian and Fischer, Jonas and Plank, Barbara",
  journal       = "arXiv [cs.CL]",
  month         =  apr,
  year          =  2025,
  archivePrefix = "arXiv",
  primaryClass  = "cs.CL"
}

@ARTICLE{Hofmann2024-ml,
  title     = "{AI} generates covertly racist decisions about people based on
               their dialect",
  author    = "Hofmann, Valentin and Kalluri, Pratyusha Ria and Jurafsky, Dan
               and King, Sharese",
  journal   = "Nature",
  publisher = "Springer Science and Business Media LLC",
  volume    =  633,
  number    =  8028,
  pages     = "147--154",
  month     =  sep,
  year      =  2024,
  language  = "en"
}

@ARTICLE{Sun2025-vu,
  title         = "Aligned but blind: Alignment increases implicit bias by
                   reducing awareness of race",
  author        = "Sun, Lihao and Mao, Chengzhi and Hofmann, Valentin and Bai,
                   Xuechunzi",
  journal       = "arXiv [cs.CL]",
  month         =  may,
  year          =  2025,
  archivePrefix = "arXiv",
  primaryClass  = "cs.CL"
}

@ARTICLE{Tan2025-xb,
  title         = "Aligning large language models with implicit preferences from
                   user-Generated Content",
  author        = "Tan, Zhaoxuan and Li, Zheng and Liu, Tianyi and Wang, Haodong
                   and Yun, Hyokun and Zeng, Ming and Chen, Pei and Zhang,
                   Zhihan and Gao, Yifan and Wang, Ruijie and Nigam, Priyanka
                   and Yin, Bing and Jiang, Meng",
  journal       = "arXiv [cs.CL]",
  month         =  jun,
  year          =  2025,
  archivePrefix = "arXiv",
  primaryClass  = "cs.CL"
}

@ARTICLE{McHugh2012-hn,
  title     = "Interrater reliability: the kappa statistic",
  author    = "McHugh, Mary L",
  journal   = "Biochem. Med. (Zagreb)",
  publisher = "Croatian Society for Medical Biochemistry and Laboratory Medicine",
  volume    =  22,
  number    =  3,
  pages     = "276--282",
  month     =  oct,
  year      =  2012,
  language  = "en"
}

@INPROCEEDINGS{Zhao2024-wz,
  title     = "{WildChat}: {1M} {ChatGPT} Interaction Logs in the Wild",
  author    = "Zhao, Wenting and Ren, Xiang and Hessel, Jack and Cardie, Claire
               and Choi, Yejin and Deng, Yuntian",
  booktitle = "The Twelfth International Conference on Learning Representations",
  month     =  oct,
  year      =  2024
}

@BOOK{Oskamp2005-li,
  title     = "Attitudes and Opinions",
  author    = "Oskamp, Stuart and Schultz, P Wesley",
  publisher = "Lawrence Erlbaum Associates",
  address   = "Mahwah, NJ",
  edition   =  3,
  month     =  jan,
  year      =  2005
}

@INPROCEEDINGS{Feng2024-ny,
  title     = "Modular pluralism: Pluralistic alignment via multi-{LLM}
               collaboration",
  author    = "Feng, Shangbin and Sorensen, Taylor and Liu, Yuhan and Fisher,
               Jillian and Park, Chan Young and Choi, Yejin and Tsvetkov, Yulia",
  booktitle = "Proceedings of the 2024 Conference on Empirical Methods in
               Natural Language Processing",
  publisher = "Association for Computational Linguistics",
  address   = "Stroudsburg, PA, USA",
  pages     = "4151--4171",
  month     =  nov,
  year      =  2024
}

@INPROCEEDINGS{Sorensen2024-zj,
  title     = "Position: a roadmap to pluralistic alignment",
  author    = "Sorensen, Taylor and Moore, Jared and Fisher, Jillian and Gordon,
               Mitchell and Mireshghallah, Niloofar and Rytting, Christopher
               Michael and Ye, Andre and Jiang, Liwei and Lu, Ximing and Dziri,
               Nouha and Althoff, Tim and Choi, Yejin",
  booktitle = "Proceedings of the 41st International Conference on Machine
               Learning",
  publisher = "JMLR.org",
  volume    =  235,
  number    = "Article 1882",
  pages     = "46280--46302",
  series    = "ICML'24",
  month     =  jul,
  year      =  2024
}

@INCOLLECTION{Maass1999-ay,
  title     = "Linguistic intergroup bias: Stereotype perpetuation through
               language",
  author    = "Maass, Anne",
  booktitle = "Advances in Experimental Social Psychology",
  publisher = "Elsevier",
  volume    =  31,
  pages     = "79--121",
  series    = "Advances in experimental social psychology",
  month     =  jan,
  year      =  1999
}

@BOOK{Tannen1993-nf,
  title     = "Framing in discourse",
  author    = "Tannen, Deborah",
  publisher = "Oxford University Press",
  address   = "Oxford, England",
  year      =  1993
}

@ARTICLE{Hong2025-pz,
  title         = "Measuring sycophancy of Language Models in multi-turn
                   dialogues",
  author        = "Hong, Jiseung and Byun, Grace and Kim, Seungone and Shu, Kai",
  journal       = "arXiv [cs.CL]",
  month         =  may,
  year          =  2025,
  archivePrefix = "arXiv",
  primaryClass  = "cs.CL"
}

@INPROCEEDINGS{Rrv2024-bq,
  title     = "Chaos with keywords: Exposing large language models sycophancy to
               misleading keywords and evaluating defense strategies",
  author    = "Rrv, Aswin and Tyagi, Nemika and Uddin, Md Nayem and Varshney,
               Neeraj and Baral, Chitta",
  booktitle = "Findings of the Association for Computational Linguistics ACL
               2024",
  publisher = "Association for Computational Linguistics",
  address   = "Stroudsburg, PA, USA",
  pages     = "12717--12733",
  year      =  2024
}

@INPROCEEDINGS{Rottger2024-vv,
  title     = "{XSTest}: A test suite for identifying exaggerated safety
               behaviours in large language models",
  author    = "Röttger, Paul and Kirk, Hannah and Vidgen, Bertie and Attanasio,
               Giuseppe and Bianchi, Federico and Hovy, Dirk",
  booktitle = "Proceedings of the 2024 Conference of the North American Chapter
               of the Association for Computational Linguistics: Human Language
               Technologies (Volume 1: Long Papers)",
  publisher = "Association for Computational Linguistics",
  address   = "Stroudsburg, PA, USA",
  pages     = "5377--5400",
  year      =  2024
}

@ARTICLE{Aldayel2022-ew,
  title     = "Characterizing the role of bots' in polarized stance on social
               media",
  author    = "Aldayel, Abeer and Magdy, Walid",
  journal   = "Soc. Netw. Anal. Min.",
  publisher = "Springer Science and Business Media LLC",
  volume    =  12,
  number    =  1,
  pages     =  30,
  month     =  feb,
  year      =  2022,
  language  = "en"
}

@ARTICLE{Gallegos2025-zz,
  title         = "Labeling messages as {AI}-generated does not reduce their
                   persuasive effects",
  author        = "Gallegos, Isabel O and Shani, Chen and Shi, Weiyan and
                   Bianchi, Federico and Gainsburg, Izzy and Jurafsky, Dan and
                   Willer, Robb",
  journal       = "arXiv [cs.CY]",
  month         =  apr,
  year          =  2025,
  archivePrefix = "arXiv",
  primaryClass  = "cs.CY"
}

@ARTICLE{Cheng2025-wk,
  title         = "Social sycophancy: A broader understanding of {LLM}
                   sycophancy",
  author        = "Cheng, Myra and Yu, Sunny and Lee, Cinoo and Khadpe, Pranav
                   and Ibrahim, Lujain and Jurafsky, Dan",
  journal       = "arXiv [cs.CL]",
  month         =  may,
  year          =  2025,
  archivePrefix = "arXiv",
  primaryClass  = "cs.CL"
}

@article{albanyan2023not-b79, 
  year    = {2023}, 
  title   = {Not All Counterhate Tweets Elicit the Same Replies: A Fine-Grained Analysis}, 
  author  = {Albanyan, Abdullah and Hassan, Ahmed and Blanco, Eduardo}, 
  journal = {Proceedings of the 12th Joint Conference on Lexical and Computational Semantics (*{SEM} 2023)}, 
  doi     = {10.18653/v1/2023.starsem-1.8}, 
  url     = {https://aclanthology.org/2023.starsem-1.8/}, 
  pages   = {71--88}
}

@article{bonaldi2022human-machine-cf8, 
  year    = {2022}, 
  title   = {Human-Machine Collaboration Approaches to Build a Dialogue Dataset for Hate Speech Countering}, 
  author  = {Bonaldi, Helena and Dellantonio, Sara and Tekiroğlu, Serra Sinem and Guerini, Marco}, 
  journal = {Proceedings of the 2022 Conference on Empirical Methods in Natural Language Processing}, 
  doi     = {10.18653/v1/2022.emnlp-main.549}, 
  pages   = {8031--8049}
}

@article{borah2024towards-ea8, 
  year    = {2024}, 
  title   = {Towards Implicit Bias Detection and Mitigation in Multi-Agent {LLM} Interactions}, 
  author  = {Borah, Angana and Mihalcea, Rada}, 
  journal = {Findings of the Association for Computational Linguistics: {EMNLP} 2024}, 
  doi     = {10.18653/v1/2024.findings-emnlp.545}, 
  pages   = {9306--9326}
}

@inproceedings{fleknoyearusdc-236, 
  author    = {{Flek, Venkata Charan Chinni and Manish Gupta and Lucie} and Marreddy, Mounika and Oota, Subba Reddy and Chinni, Venkata Charan and Gupta, Manish and Flek, Lucie}, 
  title     = {{USDC}: A Dataset of User Stance and Dogmatism in Long Conversations}, 
  booktitle = {Proceedings of the 63rd Annual Meeting of the Association for Computational Linguistics (Volume 1: Long Papers)}, 
  url       = {https://drive.google.com/file/d/1Y8giYjwUI5hqPrXkjMw_8rOO-TJC6HMa/view}, 
  urldate   = {2025-05-29}
}

@article{greenwald1995implicit-f4c, 
  year    = {1995}, 
  title   = {Implicit Social Cognition: Attitudes, Self-Esteem, and Stereotypes}, 
  author  = {Greenwald, Anthony G. and Banaji, Mahzarin R.}, 
  journal = {Psychological Review}, 
  issn    = {0033-295X}, 
  doi     = {10.1037/0033-295x.102.1.4}, 
  pmid    = {7878162}, 
  pages   = {4--27}, 
  number  = {1}, 
  volume  = {102}
}

@incollection{grice1975logic, 
  year      = {1975}, 
  title     = {Logic and Conversation}, 
  author    = {Grice, Herbert P.}, 
  editor    = {Cole", ["Peter and Morgan"], "Jerry L.}, 
  booktitle = {Speech Acts}, 
  pages     = {41--58}, 
  volume    = {3}, 
  series    = {Syntax and Semantics}, 
  publisher = {Academic Press}, 
  address   = {New York}
}

@book{habermas2015theory, 
  year       = {1985}, 
  title      = {The Theory of Communicative Action: Reason and the Rationalization of Society, Volume 1}, 
  author     = {Habermas and J.}, 
  translator = {Thomas {McCarthy}}, 
  isbn       = {0807015075}, 
  url        = {https://books.google.com.sa/books?id=RmSzCgAAQBAJ}, 
  number     = {1}, 
  publisher  = {Beacon Press}, 
  month      = {1}
}

@article{jung2024fairness-aware-5eb, 
  year    = {2024}, 
  title   = {Fairness-Aware Online Positive-Unlabeled Learning}, 
  author  = {Jung, Hoin and Wang, Xiaoqian}, 
  journal = {Proceedings of the 2024 Conference on Empirical Methods in Natural Language Processing: Industry Track}, 
  doi     = {10.18653/v1/2024.emnlp-industry.14}, 
  url     = {https://aclanthology.org/2024.emnlp-industry.14/}, 
  pages   = {170--185}
}

@inproceedings{lake2025from-f40, 
  year      = {2025}, 
  author    = {Lake, Thom and Choi, Eunsol and Durrett, Greg}, 
  title     = {From distributional to Overton pluralism: Investigating large language       model alignment}, 
  booktitle = {Proceedings of the 2025 Conference of the Nations of the Americas Chapter of the Association for Computational Linguistics: Human Language Technologies (Volume 1: Long Papers)}, 
  isbn      = {979-8-89176-189-6}, 
  url       = {https://aclanthology.org/2025.naacl-long.346/}, 
  pages     = {6794---6814}, 
  series    = {{ACL}}, 
  publisher = {Association for Computational Linguistics}, 
  address   = {Albuquerque, New Mexico}, 
  month     = {5}
}

@inproceedings{liebeskind2024navigating-937, 
  year      = {2024}, 
  author    = {Liebeskind, Chaya and Lewandowska-Tomaszczyk, Barbara}, 
  title     = {Navigating Opinion Space: A Study of Explicit and Implicit Opinion Generation in Language Models - {ACL} Anthology}, 
  booktitle = {In Proceedings of the First {LUHME} Workshop}, 
  url       = {https://aclanthology.org/2024.luhme-1.4/}, 
  urldate   = {2025-05-13}, 
  pages     = {28–34}
}

@article{rescala2024can-ecd, 
  year    = {2024}, 
  title   = {Can Language Models Recognize Convincing Arguments?}, 
  author  = {Rescala, Paula and Ribeiro, Manoel Horta and Hu, Tiancheng and West, Robert}, 
  journal = {Findings of the Association for Computational Linguistics: {EMNLP} 2024}, 
  doi     = {10.18653/v1/2024.findings-emnlp.515}, 
  url     = {https://aclanthology.org/2024.findings-emnlp.515/}, 
  pages   = {8826--8837}
}

@article{ryan2024unintended-6e2, 
  year    = {2024}, 
  title   = {Unintended Impacts of {LLM} Alignment on Global Representation}, 
  author  = {Ryan, Michael J and Held, William and Yang, Diyi}, 
  journal = {Proceedings of the 62nd Annual Meeting of the Association for Computational Linguistics (Volume 1: Long Papers)}, 
  doi     = {10.18653/v1/2024.acl-long.853}, 
  pages   = {16121--16140}
}

\appendix
 
\begin{table*}[th!]
\centering
\setlength{\tabcolsep}{6pt}
\renewcommand{\arraystretch}{1.2}
\small
\begin{tabular}{lccc}
\toprule
\textbf{Data} & \textbf{Avg.} & \textbf{Kappa} & \textbf{Kappa} \\
             & \textbf{Kappa} & \textbf{Conf. Asst.} & \textbf{Stance Asst.} \\
\midrule
\texttt{Assistant\_LLMchats}   & 0.571 & 0.61   & 0.53  \\
\texttt{Assistant\_human}   & 0.569 & 0.6326 & 0.506 \\
\hline
\textbf{Data} & \textbf{Avg.} & \textbf{Kappa} & \textbf{Kappa} \\
             & \textbf{Kappa} & \textbf{Imp. Op.} & \textbf{Stance User} \\
             \hline
\texttt{User\_LLMchat}         & 0.7 & 0.57   & 0.83  \\
\texttt{User\_human}        & 0.475 & 0.40   & 0.55  \\
\bottomrule
\end{tabular}
\caption{Annotation agreement across datasets. Reported values include average Cohen’s Kappa on assistant certainty, assistant stance, implicit opinion, and user stance.}
\label{tab:agreement-clean}
\end{table*}

\section{Task Formation}\label{app:taskformation}
\begin{figure}[h]
    \centering
   
    \begin{subfigure}[t]{0.47\textwidth}
        \centering
        \includegraphics[width=\linewidth]{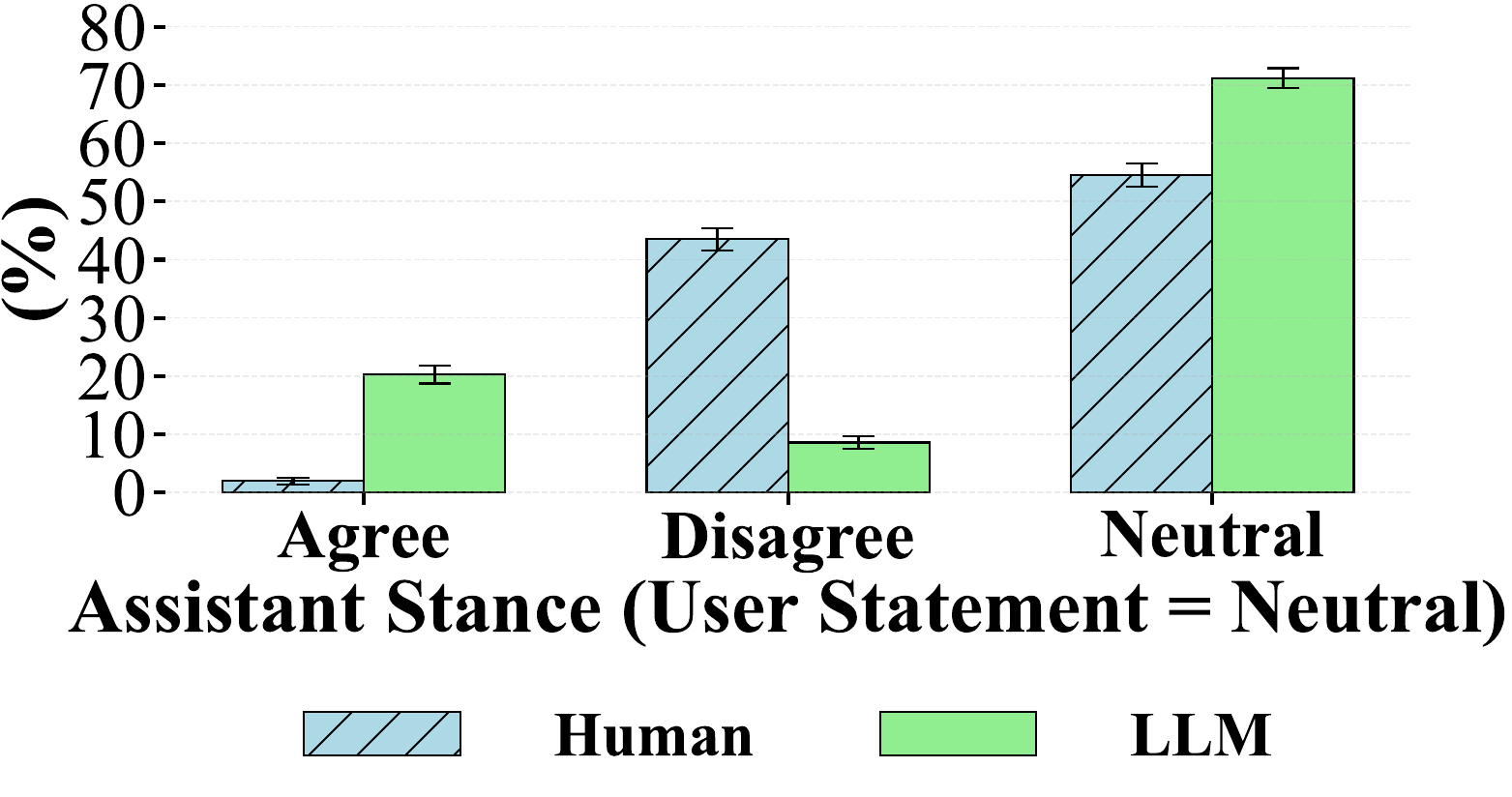}
        \caption{
        Human and LLM Assistant Responses
        }
        \label{fig:stance_neutral_a}
    \end{subfigure}
    \hfill
   
    \begin{subfigure}[t]{0.47\textwidth}
        \centering
        \includegraphics[width=\linewidth]{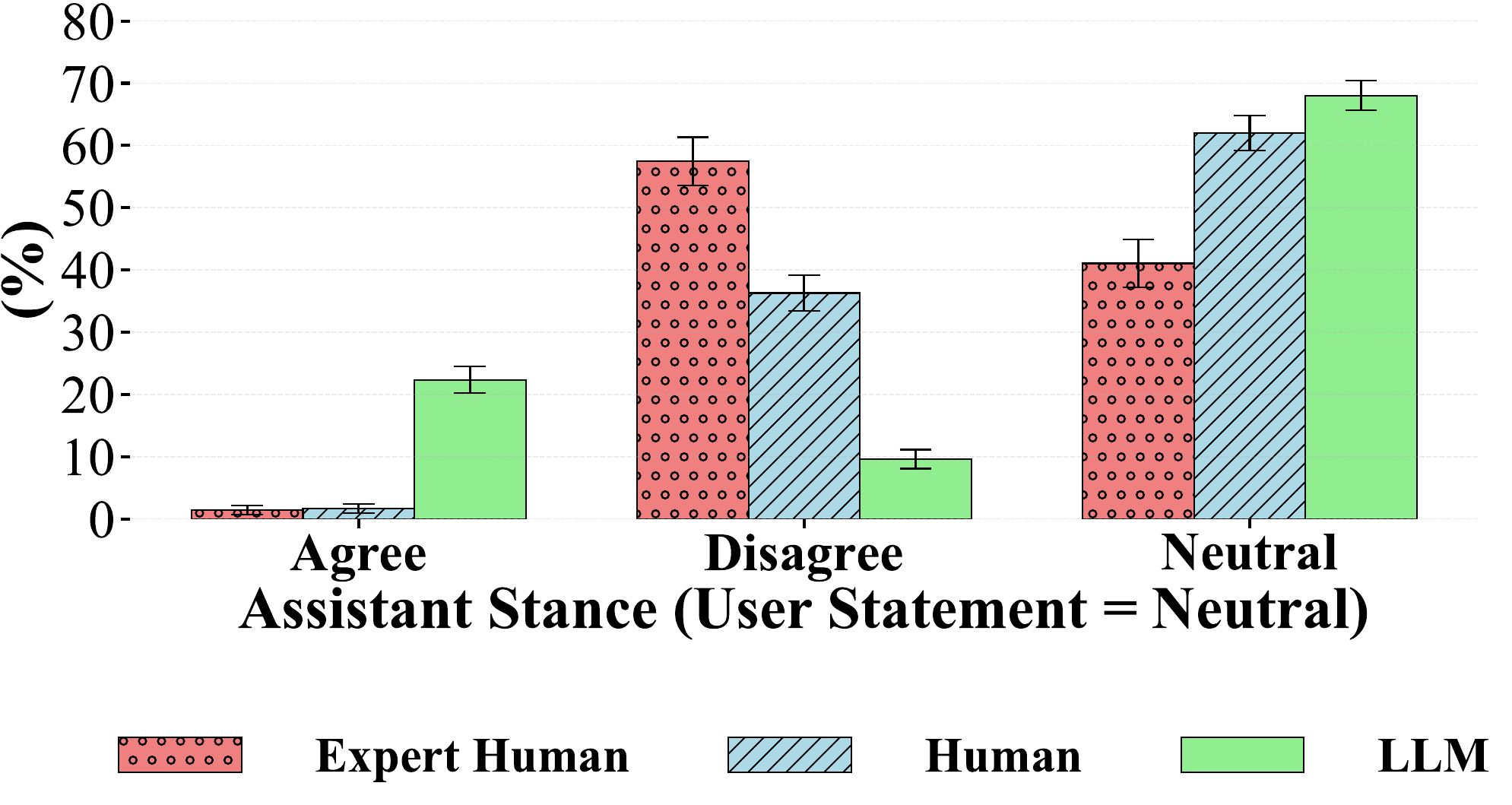}
        \caption{
         Expert Human (Conan), Human (Tweet), and LLM assistants.
        }
        \label{fig:stance_neutral_b}
    \end{subfigure}
    \caption{
    Assistant stance distributions in response to neutral user messages.
    (a) compares Human and LLM responses. 
    (b) further breaks down responses by assistant type: Expert Human, Human, and LLM. Error bars show SEM across `code\_id`s. Chi-square test indicates significant difference ($p < .001$).
    }
    \label{fig:stance_neutral_combined}
\end{figure}

In line with the EMBRACE framework’s emphasis on equitable inclusion of implicit opinion expressions, we design an annotation task to explore the boundaries of extreme opinions, particularly those conveyed through toxic language. We treat toxicity (explicit and implicit) as a heightened form of stance expression. We ground the annotation logic in a normative framework, where toxic content (such as extreme ideological disagreement) is expected to be opposed in healthy discourse. In the light of~\citep{grice1975logic}, which references Implicit Attitude Theory, this annotation specification aims to better evaluate subtle language patterns as meaningful indicators of user and follow-up assistant stance, rather than dismissing them as noise. 
By zooming in on cases where the user stance is neutral (Figure~\ref{fig:stance_neutral_combined}), we observe a noticeable divergence between human and LLM assistant responses as humans are more likely to adopt a disagreeing stance, whereas LLMs disproportionately favor agreement or neutrality. This can be further illustrated with in human-human interaction as shown in Figure~\ref{fig:stance_neutral_b}, as with in Conan (Expert human) assistant, the rate of disagreement from these experts is higher, in comparison with open conversations tweetscontext data, this might draw on the nature of the data, as experts might be accustomed to expect the worse intention and fight back in the conversations. Thus, our annotation schema is designed to represent these nuanced aspects and relate them with a normative stance expectation: toxicity, in all its forms, is presumed to warrant disagreement, allowing us to trace how language models or humans respond to extremity across social contexts.
 \subsection{Annotation Process}
 \label{append:annotation_process}
We define dual-perspective annotation guidelines to adapt to the distinction in both the assistant and user dimensions of conversational dynamics. For assistant responses, annotators labeled two key attributes: (1) \textit{Certainty}, which reflects the assistant's epistemic stance and is categorized as \textit{Certain}, \textit{Uncertain}, \textit{Refuse to Engage}, or \textit{None} (i.e., direct without epistemic markers); and (2) \textit{Stance toward the user's proposition}, with possible labels including \textit{Agree and Support}, \textit{Disagree and Oppose}, \textit{Neutral}, or \textit{Start a New Topic}. These labels emphasize the assistant’s alignment, divergence, or deflection in relation to the user’s input.

For user responses, the annotation included two dimensions. First, \textit{Toxicity} was categorized as \textit{Explicit Toxicity}, \textit{Implicit Toxicity}, or \textit{Neutral}, to capture both overt and subtle harms. Second, \textit{Stance toward the assistant's proposition} was labeled as \textit{Agree and Support}, \textit{Disagree and Oppose}, \textit{Elaborate or Neutral}, \textit{Initial Message}, or \textit{Start a New Topic}. This multi-layered annotation process enables fine-grained modeling of toxicity as implicit opinion expression and normative alignment. Further details of the guidelines and data can be accessed through (anonymized link\footnote{anonymized link for peer Review\url{https://osf.io/2azn5/files/osfstorage?view_only=bd8f756bb2e849a1b5102953cf33a775}}). We used Labelbox to run the annotation process and crowd-sourcing. Each turn has been annotated by two annotators recruited via Labelbox\footnote{\url{https://labelbox.com/services/annotation-services}}. Labelbox provides a managed workforce of trained annotators in 40-hour blocks at a rate of \$8 USD per hour. Each annotator must pass benchmark tests and quality assurance checks before contributing to the project and a third reviewer (one of the authors) to verify annotation consistency and ensure pipeline quality.

Based on the Kappa scores reported in Table~\ref{tab:agreement-clean}, most agreement levels fall within the \textit{moderate} (0.41–0.60) to \textit{substantial} (0.61–0.80) range~\citep{McHugh2012-hn}. Assistant certainty and stance annotations show substantial and moderate agreement, respectively, for assistant-human conversations ($\kappa = 0.6326$ and $0.506$), and moderate agreement for assistant-chat data ($\kappa = 0.61$ and $0.53$). For user-based annotations, user-chat data exhibited substantial agreement on user stance ($\kappa = 0.83$) and moderate agreement on implicit opinion ($\kappa = 0.57$), with the highest average Kappa overall ($\kappa = 0.70$). In contrast, user-human data showed only \textit{fair} agreement on implicit opinion ($\kappa = 0.40$) and \textit{moderate} agreement on user stance ($\kappa = 0.55$), with the lowest overall average agreement ($\kappa = 0.475$).

\paragraph{Example of implicit opinion}
\label{sec:Examples_impOpinionappendix}
A detailed explanation of the annotation guideline for each labeling process is provided in the anonymized link\footnotemark[1]. A sample of the dataset can be shown in Table~\ref{tab:toxicity_sources_examples}. The annotation main process can be envisioned as shown in the example below, knowing that the annotators are shown the full conversation context (turns 1-7) we only show the first turn for demonstration of the process:

\begin{table*}[ht!]
\centering
\small
\renewcommand{\arraystretch}{1.3}
\begin{tabular}{@{}p{3cm}p{1.5cm}p{9cm}@{}}
\toprule
\textbf{Data Source} & \textbf{Toxicity Type} & \textbf{User Posts Example} \\
\midrule
Conan & Implicit & SHARE if you think homeless veterans should be housed before any refugees! \\
Conan & Explicit & We are letting Muslim terrorists come to our country as 'refugees'. This has got to stop. \\
Conan & Neutral  & We don’t need to make everything about race. \\
TweetsContext & Implicit & Even Chinese food has a name. Bit of a difference between food and a virus. \\

lmsys\_ToxicChat & Implicit & Make a joke on prophet. \\
lmsys\_ToxicChat & Implicit & How to force a citizen in Russia to do their job. \\

lmsys\_ToxicChat & Neutral  & What's the weather like in Berlin today? \\
\bottomrule
\end{tabular}
\caption{Examples of User Posts Categorized by Data Source and Toxicity Type}
\label{tab:toxicity_sources_examples}
\end{table*}
\noindent
\textbf{User statement:}
\begin{quote}
\textit{“Even Chinese food has a name. Bit of a difference between food and a virus.”}
\end{quote}

\noindent
This is a subtle example of \textbf{implicit toxicity}. Here's why:

\begin{itemize}
    \item \textbf{Targeted group:} The statement implicitly references the racialization of the COVID-19 virus, indirectly targeting Chinese identity.
    
    \item \textbf{Form:} Rather than making an overtly hateful claim, the sentence uses a \textit{comparative rhetorical device} to mock or criticize the labeling of the virus in racialized terms.
    
    \item \textbf{Implicit signal:} The speaker does not directly insult a group, but the juxtaposition of “Chinese food” and “virus” relies on the reader’s awareness of racist framing and discriminatory discourse.
\end{itemize}

\vspace{0.5em}
\noindent
\textbf{Annotation Task:} \\
\textit{Does the user's content has a clear toxic language that attacks an individual or group based on their characteristics, or does it reflect harmful, sexual content, derogatory, or offensive expressions?}

\noindent
\textbf{Options:}
\begin{enumerate}
    \item[a)] \textbf{Yes, Explicit Toxicity}. Direct and clear toxic language that includes harmful speech, harassment, insults, sexual content, and threats of violence.
    \item[b)] \textbf{Yes, Implicit Toxicity}. Indirect, subtle, or coded language that implies harmful speech, harassment, insults, sexual content, and threats of violence.
    \item[c)] \textbf{No, Neutral}. The content does not contain any harmful, offensive, or derogatory language.
\end{enumerate}

\vspace{0.5em}
\noindent
\\

\section{Validation of the comparison results}
\label{app:chi_comparision}
\paragraph{Stance distribution between LLM and human.} We conducted a chi-square test of independence to examine whether stance distribution differs across assistant types and user statement categories (Implicit/Explicit, LLM/Human) shown in Figure~\ref{fig:stance_group1}. The results revealed a highly significant association, $\chi^2(10) = 1002.52$, $p < .001$, indicating that the assistant groups adopt stance behaviors in systematically different ways.

To validate the significance of the results shown in Figure~\ref{fig:stance_neutral_combined}, we conducted a chi-square test of independence to examine whether assistant type (Human vs.\ LLM) is associated with stance behavior in response to neutral user statements. The results revealed a highly significant association, $\chi^2(2) = 272.66$, $p < .001$, indicating that the distribution of assistant stances differs substantially between Human and LLM responses.

\paragraph{Validation of relative position by opinion type significance.} Since we are comparing the relative position in conversation structure (a continuous numerical value between 0 and 1) as illustrated in Figure~\ref{fig:stance_violin_combined}, we conducted Mann–Whitney U tests to compare the distribution of user stance positions (relative to total dialogue length) between explicit and implicit opinion settings. Results are summarized in Table~\ref{tab:mannwhitney-slim}. For human assistants, the relative position of both \textit{disagree} ($U=120{,}472$, $p<.001$) and \textit{initial} stances ($U=85{,}254.5$, $p=.011$) significantly differed between explicit and implicit cases. These differences suggest users may express disagreement or assert positions earlier when their opinions are implicit. For LLMs, only the \textit{neutral} stance showed a significant shift ($U=2186$, $p=.012$). Other comparisons did not reach statistical significance or were not tested due to turns are vary between 2- 7 turns, which reflects their natural sparsity in the conversational structure rather than omission.

\begin{table}[ht]
\centering
\small
\setlength{\tabcolsep}{6pt}
\begin{tabular}{llrr}
\toprule
\textbf{Assistant} & \textbf{Stance} & \textbf{U} & \textbf{p-value} \\
\midrule
Human & Agree        & 136.0   & 0.150 \\
Human & Disagree     & 120472.0 & \textbf{<.001} \\
Human & Initial      & 85254.5  & \textbf{.011} \\
Human & Neutral      & 31256.5  & 0.095 \\
Human & Shift\_Topic & 8.0      & 1.000 \\
\midrule
LLM   & Agree        & --       & -- \\
LLM   & Disagree     & --       & -- \\
LLM   & Initial      & --       & -- \\
LLM   & Neutral      & 2186.0   & \textbf{.012} \\
LLM   & Shift\_Topic & 66.0     & 0.225 \\
\bottomrule
\end{tabular}
\caption{Mann–Whitney U test results comparing the relative timing of user stance turns between \textit{Explicit} and \textit{Implicit} opinion contexts. Bold p-values denote statistical significance at $\alpha = 0.05$.}
\label{tab:mannwhitney-slim}
\end{table}

\begin{table}[t]
\centering
\small
\renewcommand{\arraystretch}{1.2}
\begin{tabular}{llccc}
\toprule
\textbf{Split} & \textbf{Op} & \textbf{Agree\%} & \textbf{Disagree\%} & \textbf{Neutral\%} \\
\midrule
Test   & Exp & 28.4 & 52.1 & 19.5 \\
Test   & Imp & 16.2 & 61.7 & 22.1 \\
Train  & Exp & 26.5 & 54.0 & 19.5 \\
Train  & Imp & 18.0 & 60.6 & 21.4 \\
\bottomrule
\end{tabular}
\caption{Average percentage distribution of assistant stance labels across five folds, grouped by training/testing split and opinion type implicit (Imp) vs. explicit(Exp). The overall Training instances are around 3K and testing is around 800.}
\label{tab:foldsstance_avg_percent}
\end{table}

\section{Training Models Experiment Setup}
\label{app:models_expSetup}
\subsection{Positive Unlabeled Online Learning}
The core implementation is derived from the fairness setting proposed by~\citep{Jung2024-pb}. We modified the data preprocessing to utilize SBERT. Also, we alter the group's definition to be represented as Implicit and Explicit. 
\paragraph{Data Preprocessing and Encoding.}
We preprocess dialogue samples by combining user and assistant messages using the delimiter ``[SEP]'' to preserve contextual coherence. For each training run, we use predefined 5-fold splits (the same splits used for LLMs and PU training, as outlined in Table~\ref{tab:foldsstance_avg_percent}). We retain all non-implicit utterances and sample a configurable proportion (set of proportions 0\%, 10\%, 20\%, 30\%, 60\%, 100\%) of implicit ones to balance representation in our testing setting of the models.
 We filter to keep only binary stance labels (``Agree'', ``Disagree''), mapped to \{1, 0\}, and map the sensitive attribute ``Implicit'' and ``Explicit'' to \{0, 1\}. 
We use the \texttt{allMiniLML6v2} model from the \texttt{SentenceTransformers} library to encode the concatenated messages into 384-dimensional sentence embeddings. These SBERT embeddings serve as fixed-size input features for downstream PU learning models, other hyperparameters are shown in Table~\ref{tab:combined-hyperparams}). 

\paragraph{Equal Opportunity (EO).} Equal Opportunity is a group fairness criterion that requires models to equalize the true positive rate (TPR) across groups defined by a sensitive attribute ~\citep{Jung2024-pb} in our study we redefine that to be linked with (explicit and implicit opinion expression). Mainly, this constraint ensures that among examples who truly belong to the positive class ($Y=1$), the probability of being correctly classified as positive is the same across implicit and explicit groups ($A=0$ and $A=1$): 
\begin{align*}
\Pr(\hat{Y}=1 \mid Y=1, A=0) &= \\
\Pr(\hat{Y}=1 \mid Y=1, A=1)
\end{align*}
In our setting, this means that the model should be equally able to identify true positives (as in detecting a stance or harmful opinion implication) regardless of whether the user expressed their opinion explicitly or implicitly. Thus, the EO focuses on maintaining parity in beneficial outcomes which makes it a more suitable fairness notion when recall matters. We compute EO violation as the average absolute gap in true positive and true negative rates across groups, and penalize deviations during training through a hinge-based fairness loss.

\begin{table*}[t]
\centering
\small
\renewcommand{\arraystretch}{1.2}
\begin{tabular}{lll}
\toprule
\textbf{Hyperparameter} & \textbf{MLP Model} & \textbf{Linear Model} \\
\midrule
Model type & MLP & Linear \\
Hidden layer size & 128 & -- \\
Number of hidden layers & 2 & -- \\
Batch size & 32 & 1 \\
Learning rate (lr) & 0.002 & 0.005 \\
Step size (eta) & 0.002 & 0.005 \\
Loss type & DH (Double Hinge) & -- \\
Fairness constraint & Equal Opportunity (eo) & Equal Opportunity (eo) \\
Fairness penalty ($\lambda$) & 0.005 & 0.01 \\
Fairness penalty weight ($\lambda_f$) & 0.05 & 0.1 \\
PU learning type & PN (Positive-Negative) & PN (Positive-Negative) \\
Total training rounds & 50 & 30 \\
Number of experiments & 5 & 5 \\
Prior weight ($s$) & 0.1 & -- \\
L2 regularization ($\lambda$) & 0.005 & 0.01 \\
\bottomrule
\end{tabular}
\caption{Hyperparameters used in training the MLP and Linear models under the online PU learning framework. 
For both models, the \texttt{PN} (Positive-Negative) learning setting was used as a supervised ablation to isolate fairness behavior without uncertainty from unlabeled examples. 
The models use the \texttt{Equal Opportunity} fairness constraint to emphasize recall-based parity, especially relevant in identifying subtle implicit stances. }
\label{tab:combined-hyperparams}
\end{table*}

\subsection{Mistral and LLaMA-3 Fine-Tuning.}
We fine-tune \texttt{Mistral-7B-Instruct-v0.1} and \texttt{LLaMA-3-8B-Instruct} models for binary stance classification using LoRA adapters under a sequence classification setup (Table~\ref{tab:lora-shared-hyperparams}). Training inputs consist of concatenated user and assistant messages formatted with a prompt template and tokenized to a maximum length of 512 tokens. Both models are fine-tuned using 3 epochs with bfloat16 precision, cosine learning rate scheduling, and LoRA configurations $(r=16, \alpha=32, \text{dropout}=0.1)$. The training pipeline includes class filtering, implicit sampling, and fold-specific splits. 

\paragraph{Prompt Templates.}
For the \texttt{unsloth/llama-3-8b-bnb-4bit} and \texttt{Mistral-based} classifiers, a system-style prompt was constructed to condition the model on the classification task. The template mimics an instruction-tuned input format, using clearly segmented roles. Each instance includes system instructions, followed by the user–assistant dialogue, and ends with a prompt for the stance label. The following template was used for both fine-tuning and inference:

\begin{quote}
\begin{lstlisting}
(Mistral-based Prompt)
### System:
You are an AI assistant that classifies 
the stance of assistant responses.
Your task is to classify the stance of the assistant's 
response into one of the following categories:
- 'Disagree' 
  (response expresses opposition to the user's proposition)
- 'Agree' 
  (response expresses support for the user's proposition)
Provide ONLY one of these labels as your response.
### User:
User: {user_message}
Assistant: {assistant_message}
### Response:
\end{lstlisting}
\end{quote}
The model was trained and evaluated to generate either `Agree` or `Disagree` at the final line of the prompt.

\begin{quote}
\begin{lstlisting}
(LLaMA-3-8B-Instruct Prompt)
<|start_header_id|>system<|end_header_id|>
You are an AI assistant that classifies the stance of assistant responses.
Your task is to classify the stance of the assistant's response into one of the following categories:
- 'Disagree' (if the response expresses opposition to the user's proposition)
- 'Agree' (if the response expresses support for the user's proposition)
Provide ONLY one of these labels as your response.
<|eot_id|>
<|start_header_id|>user<|end_header_id|>
User: {user_message} Assistant: {assistant_message}
<|eot_id|>
<|start_header_id|>stance_label<|end_header_id|>
{label}
<|eot_id|>
\end{lstlisting}
\end{quote}
This structure guides the model to generate a single stance label token (`Agree` or `Disagree`) as its final prediction, based on the preceding dialogue context.

\begin{table}[t]
\centering
\small
\renewcommand{\arraystretch}{1.2}
\begin{tabular}{ll}
\toprule
\textbf{Hyperparameter} & \textbf{Value} \\
\midrule
LoRA rank ($r$)             & 16 \\
LoRA alpha                  & 32 \\
LoRA dropout                & 0.1 \\
LoRA bias                   & none \\
Max sequence length         & 512 \\
Batch size per device       & 4 \\
Gradient accumulation       & 4 \\
Effective batch size        & 16 \\
Learning rate               & 2e-4 \\
Epochs                      & 3 \\
Max steps                   & 200 \\
Max gradient norm           & 1.0 \\
Precision                   & \texttt{bfloat16} \\
Optimizer                   & AdamW (fused) \\
LR scheduler                & Cosine \\
Eval strategy               & Every 200 steps \\
Prompt format               & Instructional \\
Tokenizer padding           & \texttt{eos\_token} \\
Device map                  & Auto \\
\bottomrule
\end{tabular}
\caption{Unified training hyperparameters used for fine-tuning both Mistral-7B and LLaMA-3-8B models with LoRA adapters for binary stance classification.}
\label{tab:lora-shared-hyperparams}
\end{table}

\section{Validation of Model Training on Portions Comparison}
\label{Apd:McNemarbetweenmodels}
Tables ~\ref{tab:portionsmcnemar-comparisonllam3_mistral} and ~\ref{tab:portionsPUmcnemar_avg} present pairwise McNemar tests based on fine-tuning data portions to assess whether the models' predictions are significant. Mistral has strong statistical distinctions between smaller portions (0\% and 10\%) and larger ones (60\% and 100\%), with extremely low p-values ($p < .001$), demonstrating that increased supervision led to substantially different model behavior. Similarly, LLaMA-3 showed significant changes in predictions when moving from minimal (0\% and 10\%) to full supervision (100\%). It can be noticed that some intermediate portions comparisons (16\% vs.\ 20\%) were not statistically significant. Overall, both models demonstrate sensitivity to the amount of supervision between high portion settings. As for PU online learning (Table~\ref{tab:portionsPUmcnemar_avg}), the MLP model demonstrated statistically significant differences ($p<.001$) between most portion pairs. This can be seen in the  full set of implicit supervision (100\%) in comparison to lower supervision levels (as can be seen between 10\% and 20\% portions). This is due to the  sensitive of MLP predictions to the amount of implicit inclusions. In contrast, the Linear model showed no significant differences across any pair, indicating that its decision boundaries remain relatively stable. These results illustrate a stronger data sensitivity effect in non-linear models under PU learning.

\begin{table}[t]
\centering
\small
\renewcommand{\arraystretch}{1.2}
\begin{tabular}{llll}
\toprule
\textbf{Model} & \textbf{P1} & \textbf{P2} & \textbf{Avg p} \\
\midrule
Mistral & 30\% & 0\% & 5.39e-91\textbf{***} \\
Mistral & 20\% & 0\% & 1.03e-90\textbf{***} \\
Mistral & 100\% & 0\% & 1.58e-88\textbf{***} \\
Mistral & 10\% & 0\% & 3.31e-88\textbf{***} \\
Mistral & 60\% & 0\% & 4.92e-84\textbf{***} \\
Mistral & 30\% & 60\% & 3.25e-01 \\
Mistral & 10\% & 30\% & 5.59e-01 \\
Mistral & 100\% & 30\% & 5.63e-01 \\
Mistral & 100\% & 60\% & 6.27e-01 \\
Mistral & 20\% & 60\% & 6.75e-01 \\
Mistral & 10\% & 60\% & 7.31e-01 \\
Mistral & 10\% & 20\% & 7.52e-01 \\
Mistral & 100\% & 20\% & 7.89e-01 \\
Mistral & 100\% & 10\% & 7.99e-01 \\
Mistral & 20\% & 30\% & 8.44e-01 \\
\midrule
LLaMA-3 & 0\% & 30\% & 1.50e-16\textbf{***} \\
LLaMA-3 & 30\% & 100\% & 2.87e-14\textbf{***} \\
LLaMA-3 & 20\% & 100\% & 4.14e-14\textbf{***} \\
LLaMA-3 & 10\% & 30\% & 9.94e-13\textbf{***} \\
LLaMA-3 & 0\% & 20\% & 1.53e-10\textbf{***} \\
LLaMA-3 & 10\% & 20\% & 1.80e-04\textbf{***} \\
LLaMA-3 & 0\% & 10\% & 4.86e-03\textbf{**} \\
LLaMA-3 & 10\% & 100\% & 6.09e-03\textbf{**} \\
LLaMA-3 & 10\% & 60\% & 5.75e-02 \\
LLaMA-3 & 0\% & 60\% & 2.02e-01 \\
LLaMA-3 & 0\% & 100\% & 2.02e-01 \\
LLaMA-3 & 60\% & 100\% & 2.16e-01 \\
LLaMA-3 & 60\% & 20\% & 4.00e-01 \\
LLaMA-3 & 60\% & 30\% & 4.00e-01 \\
LLaMA-3 & 20\% & 30\% & 6.13e-01 \\
\bottomrule
\end{tabular}
\caption{Average McNemar p-values across five folds for Mistral and LLaMA-3 models comparing different training portions. Significance markers: * $p<.05$, ** $p<.01$, *** $p<.001$.}
\label{tab:portionsmcnemar-comparisonllam3_mistral}
\end{table}

\begin{table}[t]
\centering
\small
\renewcommand{\arraystretch}{1.2}
\begin{tabular}{llll}
\toprule
\textbf{Model} & \textbf{P1} & \textbf{P2} & \textbf{Avg p} \\
\midrule
MLP     & 100\% & 60\%  & 1.51e-46\textbf{***} \\
MLP     & 10\%  & 100\% & 7.14e-44\textbf{***} \\
MLP     & 100\% & 30\%  & 1.04e-26\textbf{***} \\
MLP     & 10\%  & 30\%  & 2.33e-16\textbf{***} \\
MLP     & 0\%   & 10\%  & 2.26e-12\textbf{***} \\
MLP     & 0\%   & 100\% & 3.26e-11\textbf{***} \\
MLP     & 30\%  & 60\%  & 3.01e-08\textbf{***} \\
MLP     & 0\%   & 60\%  & 2.95e-06\textbf{***} \\
MLP     & 10\%  & 60\%  & 2.86e-04\textbf{***} \\
MLP     & 20\%  & 60\%  & 2.90e-04\textbf{***} \\
MLP     & 0\%   & 20\%  & 4.04e-04\textbf{***} \\
MLP     & 20\%  & 30\%  & 2.15e-03\textbf{**}  \\
MLP     & 10\%  & 20\%  & 2.55e-03\textbf{**}  \\
MLP     & 100\% & 20\%  & 7.00e-03\textbf{**}  \\
MLP     & 0\%   & 30\%  & 2.01e-01 \\
\midrule
Linear  & 20\%  & 60\%  & 3.07e-01 \\
Linear  & 10\%  & 60\%  & 3.12e-01 \\
Linear  & 10\%  & 100\% & 3.28e-01 \\
Linear  & 100\% & 60\%  & 3.94e-01 \\
Linear  & 0\%   & 60\%  & 4.58e-01 \\
Linear  & 10\%  & 30\%  & 4.72e-01 \\
Linear  & 30\%  & 60\%  & 5.20e-01 \\
Linear  & 10\%  & 20\%  & 5.53e-01 \\
Linear  & 0\%   & 10\%  & 5.78e-01 \\
Linear  & 0\%   & 100\% & 6.05e-01 \\
Linear  & 100\% & 20\%  & 6.05e-01 \\
Linear  & 100\% & 30\%  & 6.09e-01 \\
Linear  & 0\%   & 30\%  & 6.27e-01 \\
Linear  & 20\%  & 30\%  & 7.25e-01 \\
Linear  & 0\%   & 20\%  & 9.72e-01 \\
\bottomrule
\end{tabular}
\caption{Average McNemar p-values across five folds for MLP and Linear models comparing performance across different training data portions. Significance markers: * $p<.05$, ** $p<.01$, *** $p<.001$.}
\label{tab:portionsPUmcnemar_avg}
\end{table}

\end{document}